\documentclass[letterpaper, 10 pt, journal, twoside]{IEEEtran} 

\IEEEoverridecommandlockouts                              

\usepackage[dvipsnames]{xcolor}
\definecolor{dark-green}{RGB}{12,80,12}

\usepackage{url}
\usepackage{dsfont}
\usepackage{amsmath}
\usepackage{amssymb}
\usepackage{cuted}
\usepackage{multirow} 
\usepackage{multicol}
\usepackage{flushend}
\usepackage[shortlabels]{enumitem}

\usepackage{tabularx}
\newcolumntype{C}[1]{>{\centering\let\newline\\\arraybackslash\hspace{0pt}}m{#1}} 
\newcolumntype{L}[1]{>{\let\newline\\\arraybackslash\hspace{0pt}}m{#1}} 

\usepackage[breaklinks,colorlinks]{hyperref}
\usepackage[mathscr]{euscript}

\usepackage[export]{adjustbox}
\newcolumntype{P}[1]{>{\centering\arraybackslash}p{#1}}

\iffalse 
  \newcommand{\todo}[1]{\noindent}
\else
  \newcommand{\todo}[1]{\textcolor{red}{\bf [Todo: #1]}}
  
\fi

\usepackage{booktabs}
\usepackage{xspace}
\usepackage{caption}
\usepackage{graphicx} 
\usepackage{color}
\usepackage{siunitx}
\usepackage{subcaption}
\usepackage[hang,flushmargin,symbol]{footmisc}
\renewcommand{\thefootnote}{\fnsymbol{footnote}}
\usepackage{microtype}
\usepackage{rotating}
\newcommand{\secref}[1]{Sec.~\ref{#1}}
\renewcommand{\eqref}[1]{Eq.~(\ref{#1})}
\newcommand{\figref}[1]{Fig.~\ref{#1}}
\newcommand{\tabref}[1]{Tab.~\ref{#1}}

\usepackage{tabularx}
\usepackage{colortbl} 
\usepackage{array, makecell} %
\newcolumntype{Y}{>{\centering\arraybackslash}X}
\newcolumntype{Z}{>{\raggedleft\arraybackslash}X}

\usepackage{array}
\usepackage{multirow}
\usepackage{pifont}
\usepackage{cite}
\usepackage[export]{adjustbox}
\usepackage[resetlabels]{multibib}
\newcites{New}{References}

\usepackage{tikz}

\iffalse 
  \newcommand{\mohan}[1]{\noindent}
  \newcommand{\cattaneo}[1]{\noindent}
  \newcommand{\arce}[1]{\noindent}
  \newcommand{\abhi}[1]{\noindent}
  \newcommand{\todo}[1]{\noindent}
\else
  \newcommand{\mohan}[1]{\textcolor{blue}{\bf [RM: #1]}}
  \newcommand{\cattaneo}[1]{\textcolor{orange}{\bf [DC: #1]}}
  \newcommand{\arce}[1]{\textcolor{purple}{\bf [JA: #1]}}
  \newcommand{\abhi}[1]{\textcolor{green}{\bf [AV: #1]}}
\fi

\DeclareSIUnit{\rad}{rad}

\renewcommand{\baselinestretch}{0.99}

\newcommand{\datasetname}{{\mbox{Syn-Mediverse}}}

\title{\LARGE \bf
\datasetname{}: A Multimodal Synthetic Dataset for Intelligent Scene Understanding of Healthcare Facilities
}

\author{Rohit Mohan, José Arce, Sassan Mokhtar, Daniele Cattaneo, Abhinav Valada
\thanks{Department of Computer Science, University of Freiburg, Germany.}%
}

\begin{document}

\maketitle
\thispagestyle{empty}
\pagestyle{empty}

\begin{abstract}
Safety and efficiency are paramount in healthcare facilities where the lives of patients are at stake. Despite the adoption of robots to assist medical staff in challenging tasks such as complex surgeries, human expertise is still indispensable. The next generation of autonomous healthcare robots hinges on their capacity to perceive and understand their complex and frenetic environments. While deep learning models are increasingly used for this purpose, they require extensive annotated training data which is impractical to obtain in real-world healthcare settings. To bridge this gap, we present \datasetname{}, the first hyper-realistic multimodal synthetic dataset of diverse healthcare facilities. \datasetname{} contains over \num{48000} images from a simulated industry-standard optical tracking camera and provides more than 1.5M annotations spanning five different scene understanding tasks including depth estimation, object detection, semantic segmentation, instance segmentation, and panoptic segmentation. We demonstrate the complexity of our dataset by evaluating the performance on a broad range of state-of-the-art baselines for each task. To further advance research on scene understanding of healthcare facilities, along with the public dataset we provide an online evaluation benchmark available at \url{http://syn-mediverse.cs.uni-freiburg.de}.
 
\end{abstract}

\section{Introduction}

Intelligent scene understanding of healthcare facilities has gained significant interest in recent years. This is primarily due to the increasing number of medical robotics systems such as surgical robots in operating rooms and autonomous supplies delivery robots. An accurate scene understanding system can help improve the efficiency of surgical procedures, for example, by aiding surgical auditing via automated or semi-automated report generation. Moreover, automatic scene understanding can facilitate a higher level of autonomy in the next generation of medical robots, thereby reducing time and errors in hospitals, which are one of the most complex, and exigent workplaces.

\begin{figure}
    \centering
        \includegraphics[width=\linewidth]{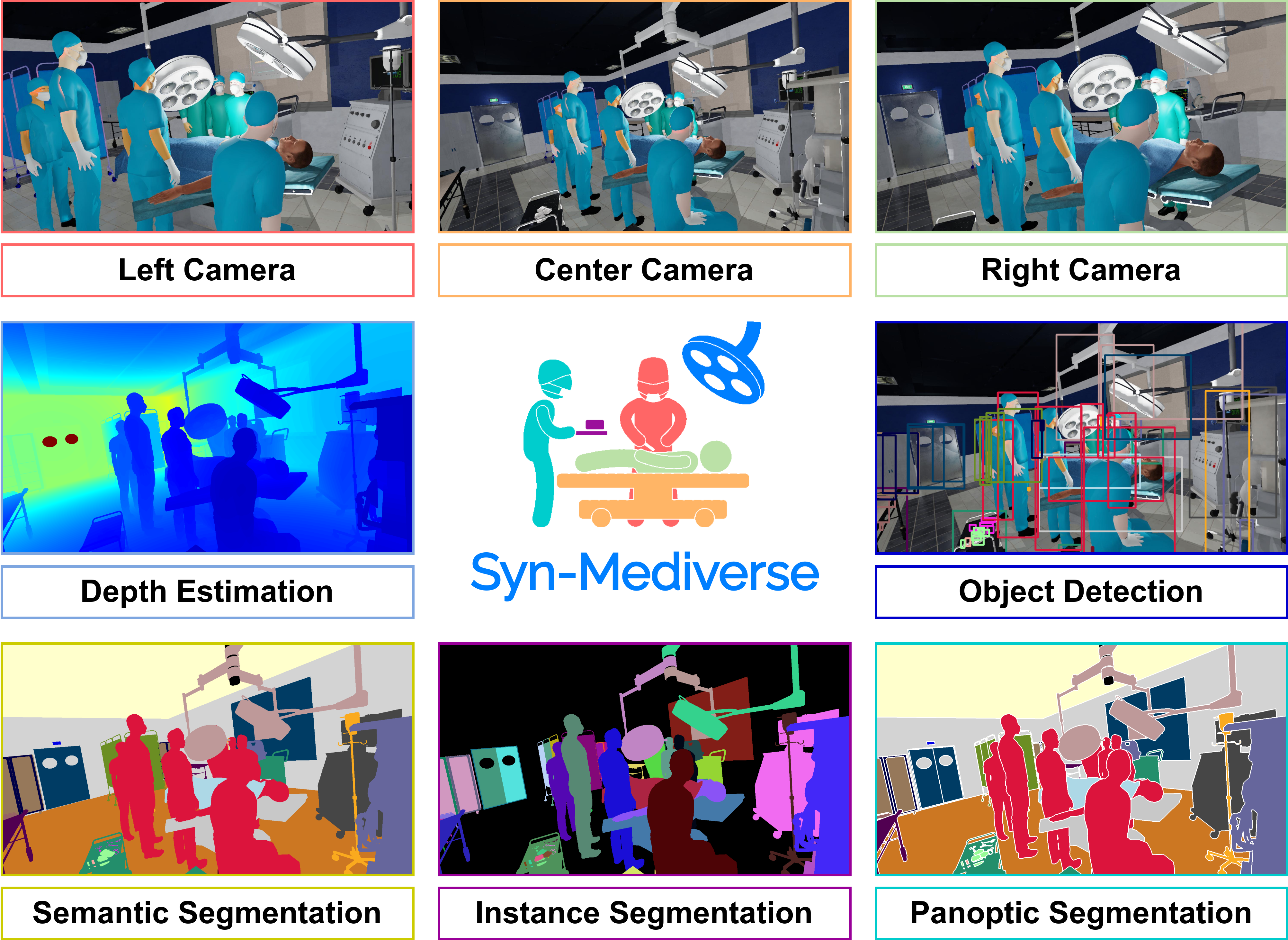}
    \caption{Overview of the \datasetname{} dataset consisting of multi-view images captured from a simulated industry-standard optical tracking camera. We provide pixel-level annotations for depth estimation, object detection, semantic segmentation, instance segmentation, and panoptic segmentation.}
    \label{fig:teaser}
\end{figure}

Fundamental perception tasks such as object detection~\cite{lang2022robust}, panoptic segmentation~\cite{mohan2020amodal}, and depth estimation~\cite{vodisch2023codeps} have been extensively studied in the context of everyday environments such as indoor houses, office spaces, and outdoor urban scenes, with several related datasets and benchmarks publicly available~\cite{hurtado2022semantic}. Medical facilities, however, present a different set of challenges when compared to these environments. For instance, many objects present in an operating room are unique to the healthcare setting, and facial and body features are typically occluded behind surgical masks and other hospital gear. Hence, existing approaches developed for everyday environments do not directly generalize well to the healthcare domain.

Collecting and annotating data from real-world healthcare facilities poses significant challenges due to the fast-paced nature of hospital activities, strict sterility rules, and stringent privacy protection regulations. Consequently, existing datasets in healthcare settings are typically tailored towards a single task such as object detection~\cite{fujii2022surgicaltools}, semantic segmentation~\cite{colleoni2020exvivo,grammatikopoulou2021cadis,li2020orenvaware}, or human pose estimation~\cite{srivastav2021mvor,belagiannis2016skeletons}, and are often confined to a single room, limiting their diversity. Moreover, accessibility to some of these datasets is restricted or not publicly available due to privacy concerns~\cite{li2020orenvaware, sharghi2020autoor}, making it furthermore challenging to advance research in this domain.

In this paper, we propose \datasetname{}, a first-of-a-kind multimodal multitask scene understanding dataset of healthcare facilities. Our novel dataset is composed of RGB-D images captured in varying illumination conditions and from a wide variety of scenes including 13 different rooms such as operating rooms and consultation rooms with different types of medical equipment including surgical robots and surgical tools. We provide ground truth annotations for five different tasks: object detection, semantic segmentation, instance segmentation, panoptic segmentation, and monocular depth estimation. \figref{fig:teaser} presents an overview of the dataset. \datasetname{} contains over 1.5M annotations spanning 31 distinct semantic classes. This is the first publicly available dataset for scene understanding of diverse healthcare environments with surgical robots. Moreover, the dataset presents an opportunity to directly evaluate on real-world scenes using transfer learning. Additionally, we maintain an online evaluation benchmark\footnote{The benchmark server will be public upon acceptance.} for the five fundamental perception tasks and provide a wide variety of state-of-the-art baselines for each of them. The benchmark and dataset are available at \url{http://syn-mediverse.cs.uni-freiburg.de}.

\newcommand*{\chk}[1][]{\tikz[x=1em, y=1em]\fill[#1] (0,.35) -- (.25,0) -- (1,.7) -- (.25,.2) -- cycle;}
\newcommand{\STAB}[2][c]{\begin{tabular}{@{}#1@{}}#2\end{tabular}}
\newcommand*{\vertical}[1]{\begin{sideways}{#1}\end{sideways}}
\newcommand*{\ua}{\textbf{*}}

\begin{table*}[ht]
    \centering
    \captionsetup{width=.70\linewidth}
    \caption{Comparison of different medical datasets, grouped by the scope of inputs. The data from datasets marked with \ua{} was neither publicly available nor easily accessible at the time of this writing.}
    \label{tab:datasets}
    
    \footnotesize
    \begin{tabular}{clcc|ccc|cc|cccr}
    \toprule
    & & \multicolumn{9}{c}{Task / Annotations} &  \\
    \cmidrule(lr){3-11}
    & & & & \multicolumn{3}{c}{Segmentation} & \multicolumn{2}{c}{Depth} & & & & \multirow{2}{*}[-5.3em]{\makecell{Annotated\\Samples}}\\
    \cmidrule(lr){5-7} \cmidrule(lr){8-9}
    & Dataset & \vertical{Classification} & \vertical{Detection} & \vertical{Instance} & \vertical{Semantic} & \vertical{Panoptic} & \vertical{Monocular} & \vertical{Stereo} & \vertical{Human Pose} & \vertical{Scene Graph} & Type \\
    \midrule
    
    \multirow{4}{*}{\vertical{MIS}}
    & Cholect50~\cite{nwoye2022cholect50}               & \chk &   -  &   -  &  -   &  -   &  -   &  -   &   -  &   -  & Real      & \num{100863} \\
    & DSAD~\cite{carstens2023dsad}                      & \chk &   -  &  -   & \chk &  -   &  -   &  -   &   -  &   -  & Real      & \num{ 14625} \\
    & AutoLaparo~\cite{wang2022autolaparo}              & \chk &   -  & \chk &  -   &  -   &  -   &  -   &   -  &   -  & Real      & \num{  1800} \\
    & HeiCo~\cite{maier2021heico}                       & \chk &   -  & \chk &  -   &  -   &  -   &  -   &   -  &   -  & Real      & \num{ 10040} \\
    
    \midrule
    
    \multirow{3}{*}{\vertical{\makecell{Open\\Surgery}}}
    & CATARACTS~\cite{al2019cataracts}                  & \chk &   -  &  -   &   -  &  -   &  -   &  -   &   -  &  -   & Real      & \num{957884} \\
    & Surgical Tools~\cite{fujii2022surgicaltools}\ua{} &  -   & \chk &  -   &   -  &  -   &  -   &  -   &   -  &  -   & Real      & \num{ 19560} \\
    & CaDIS~\cite{grammatikopoulou2021cadis}            &  -   &   -  &  -   & \chk &  -   &  -   &  -   &   -  &  -   & Real      & \num{  4670} \\
    
    \midrule
    
    \multirow{6}{*}[-0.5em]{\vertical{General OR}}
    & Auto OR~\cite{sharghi2020autoor}\ua{}             & \chk &   -  &   -  &  -   &   -  &   -  &  -   &  -   &   -  & Real      & N/A \\
    & MVOR~\cite{srivastav2021mvor}                     &  -   & \chk &   -  &  -   &   -  & \chk &  -   & \chk &   -  & Real      & \num{   732} \\
    & Robotic3DOR~\cite{li2020orenvaware}\ua{}          &  -   &   -  &   -  & \chk &   -  &   -  &  -   &   -  &   -  & Simulated & \num{  7980} \\
    & MultiHumanOR~\cite{belagiannis2016skeletons}      &  -   &   -  &   -  &  -   &   -  &   -  &  -   & \chk &   -  & Simulated & \num{   700} \\
    & 4D-OR~\cite{oszoy20224dor}                        &  -   & \chk &   -  &  -   &   -  & \chk & \chk & \chk & \chk & Simulated & \num{  6780} \\
    \cmidrule{2-13}
    & \textbf{\datasetname{} (Ours)}                    &  -   & \chk & \chk & \chk & \chk & \chk &   -  &  -   &   -  & Synthetic & \num{48000} \\
    
    \bottomrule
\end{tabular}
\end{table*}

\section{Related Work}
\label{sec:relatedWork}

In this section, we review some of the most commonly used datasets for different perception tasks in the field of healthcare. The datasets are grouped by the scope and level of detail that they try to capture into three categories, namely: laparoscopic or minimally invasive surgeries (MIS), open surgeries, and general operating room (OR). It is in this last category that our proposed \datasetname{} falls into but since there are only a handful of related publicly available datasets, which focus mostly on pose estimation, we include the other categories for completeness.  In \tabref{tab:datasets}, we present a comparison of the different datasets, the available annotations, and their sizes in terms of the number of samples.

{\parskip=5pt\noindent\textit{Minimally Invasive Surgery}: } These datasets typically contain single-view RGB images captured with an endoscopic or laparoscopic camera, and at an extreme close-up angle, differing in the type of procedures and available annotations. The Cholect50~\cite{nwoye2022cholect50} dataset consists of 50 videos of real laparoscopic colon surgeries, annotated for the task of classifying triplets of $\langle$instrument, action, target tissue$\rangle$ from sequences of frames. Similarly, the Dresden Surgical Anatomy Dataset~\cite{carstens2023dsad} is comprised of 32 videos from robot-assisted colon surgeries, with semantic segmentation masks for the different types of tissues and multilabel class annotations per frame to detect the presence of organs and tissues. AutoLaparo~\cite{wang2022autolaparo} consists of 30 videos of three types of colon procedures that include segmentation masks for the instruments, as well as captions for sequences of frames that are used to determine the current phase of the surgery. 


{\parskip=5pt\noindent\textit{Open Surgery}:} These datasets also typically contain single-view RGB images but are captured from a dedicated navigation or head-mounted camera at a relatively close angle focused on the areas being operated on. CATARACTS~\cite{al2019cataracts} consists of images from 100 eye surgery videos, whose labels are used to identify the presence of surgical instruments in a given frame. CaDIS~\cite{grammatikopoulou2021cadis} uses the same input images from CATARACTS but adds semantic segmentation masks for the tissues and instruments. The Surgical Tools~\cite{fujii2022surgicaltools} dataset provides bounding boxes for 31 types of instruments from 15 plastic surgeries, captured with a camera worn by the surgeons on their heads. 



{\parskip=5pt\noindent\textit{General OR}:} The datasets presented in this section consist of multi-view RGB or range images obtained from a wider angle, capturing the entire scene of an operating room. The objective of most datasets in this category is to provide scene understanding and situational awareness in the operating room. For instance, both the MVOR~\cite{srivastav2021mvor} and MultiHumanOR~\cite{belagiannis2016skeletons} datasets contain labels for predicting the keypoints and poses of the upper part of the bodies of the patient and medical staff. AutoOR~\cite{sharghi2020autoor} consists of sequences of range images, captured using multiple Time-of-Flight (TOF) cameras from 103 actual surgeries and labeled to predict the current phase of the operation. Robotic3DOR~\cite{li2020orenvaware} includes semantic segmentation masks for simulated laparoscopic surgeries, captured with multiple TOF cameras. Finally, 4D-OR~\cite{oszoy20224dor} includes aggregated point clouds, 3D bounding boxes, 6 degrees-of-freedom (DoF) human poses, and semantic scene graphs from simulated total knee replacement surgeries. 

Syn-Mediverse sets itself apart by offering an extensive and diverse understanding of medical facilities ranging from operating rooms to consultation rooms. This diversity is further enriched by the inclusion of advanced assistance systems such as surgical robots. Given that our dataset offers annotations for five fundamental perception tasks, features that are notably lacking in previously mentioned datasets, Syn-Mediverse stands as a valuable resource, promising to not only complement but also elevate the existing body of medical datasets.

\section{\datasetname{} Dataset}
\label{sec:dataset}

In this section, we first describe the dataset creation procedure in~\secref{subsec:creation}, followed by an outline of the dataset structure in~\secref{subsec:struct}. We then detail the benchmarking protocol in~\secref{subsec:benchmark} and present a detailed statistical analysis of our dataset in~\secref{subsec:stats}.

\begin{figure}[t]
    \centering
    \begin{subfigure}[b]{\linewidth}
        \centering
        \includegraphics[width=\textwidth]{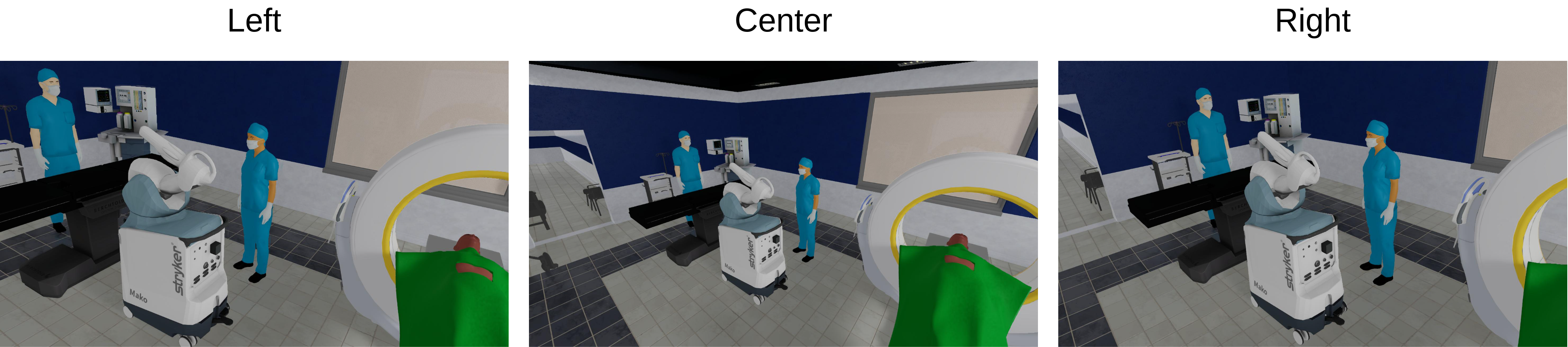} 
        \subcaption{Scene captured by the multi-camera setup in the \datasetname{} dataset.}
        \label{FP8000_fig}
        \vspace*{0.2cm}%
    \end{subfigure}
    \begin{subfigure}[b]{0.49\linewidth}
        \centering
        \includegraphics[width=\textwidth]{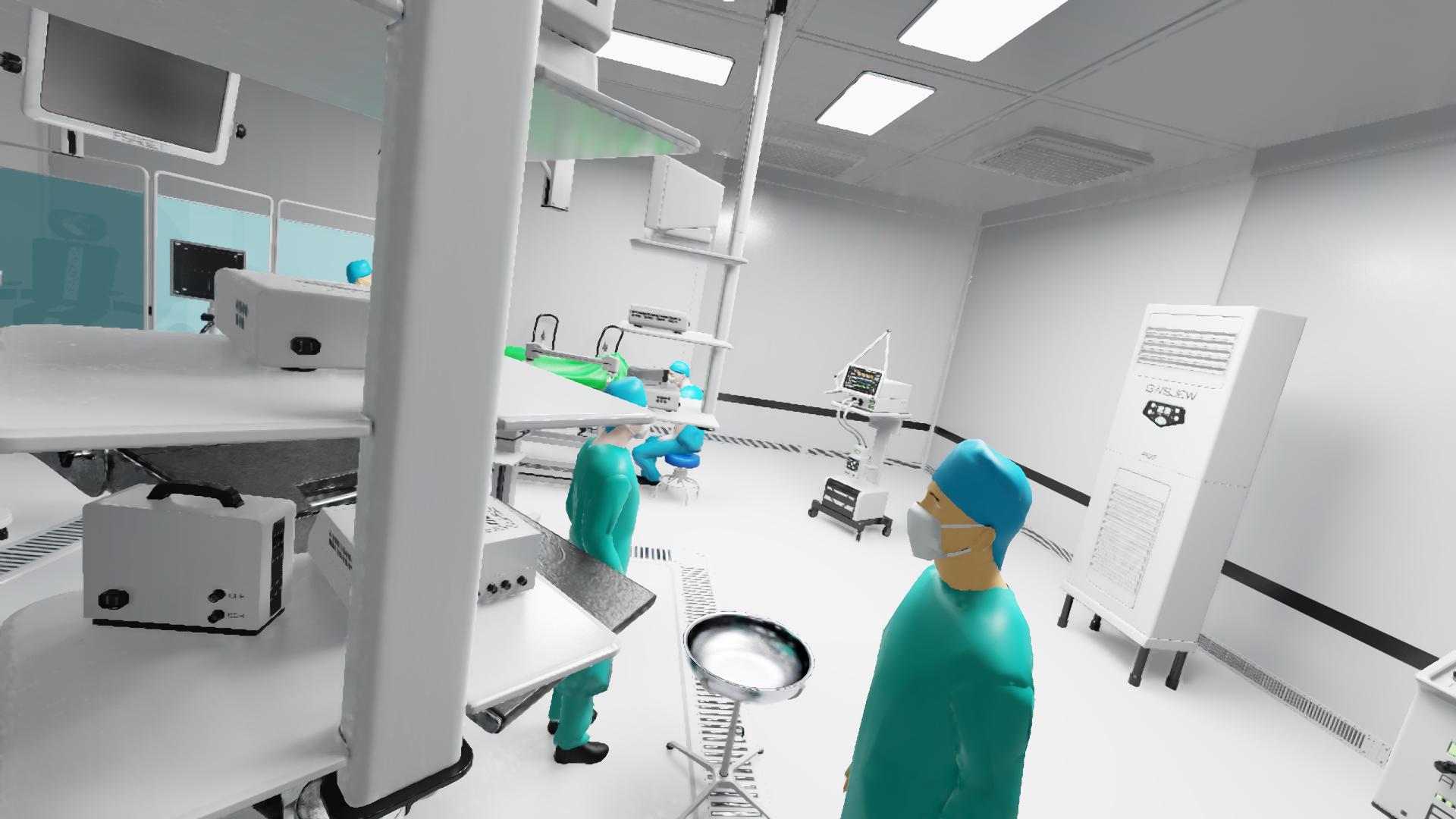}
        \subcaption{Initial synthetic image.}
        \label{init}
    \end{subfigure}
    \begin{subfigure}[b]{0.49\linewidth}
        \centering
        \includegraphics[width=\textwidth]{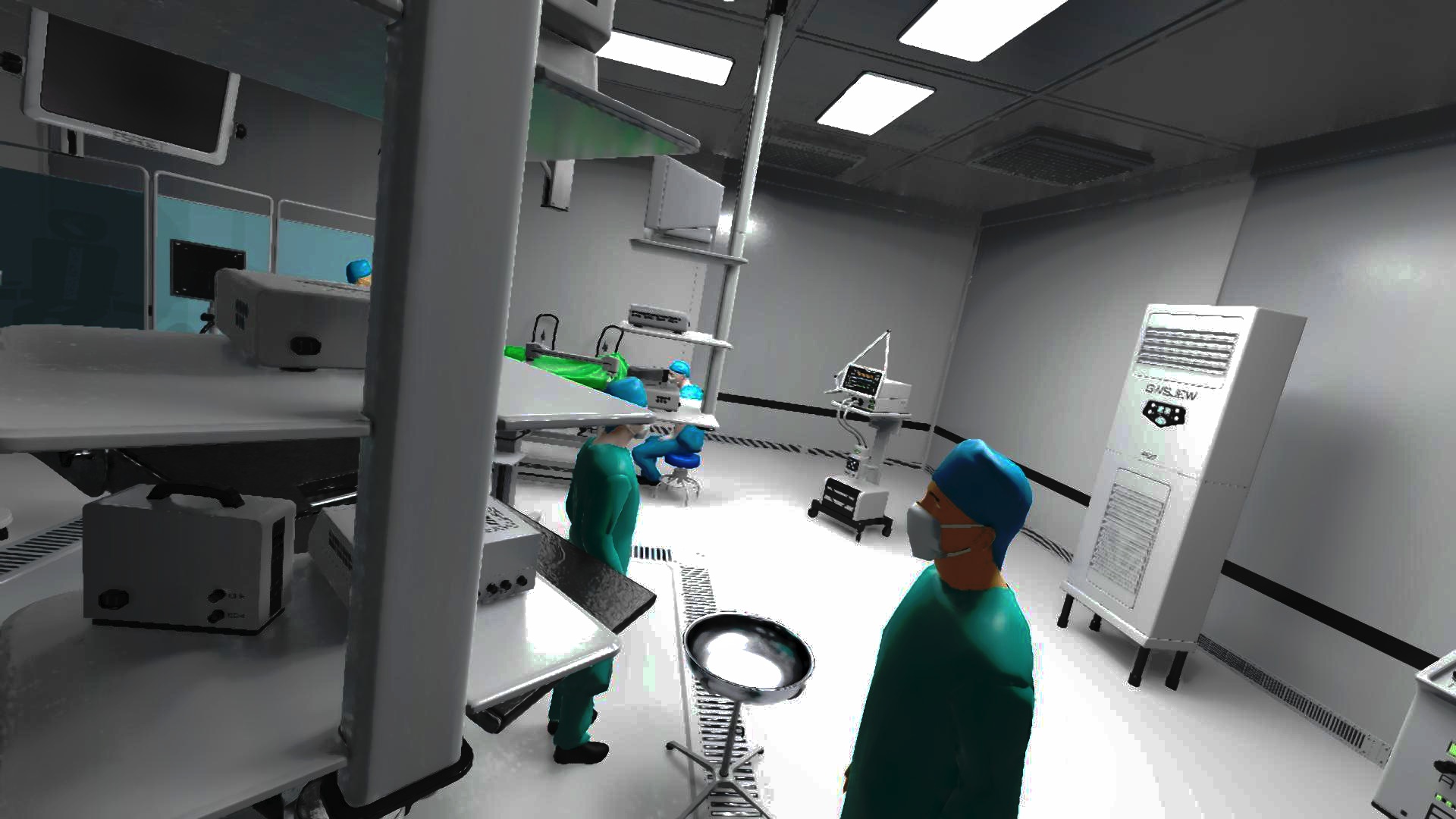}
        \subcaption{Synthetic image after adding noise.}
        \label{with_noise}
    \end{subfigure}
    \caption{Depiction of the scene captured with the multi-camera setup in (a), and illustration of a synthetic image pre- and post-noise addition via histogram matching in (b) and (c).}
    \label{fig:stats-ood}
\end{figure}

\begin{figure*}
\centering
\footnotesize
\setlength{\tabcolsep}{0.1cm}
\begin{tabular}{P{2.8cm}P{2.8cm}P{2.8cm}P{2.8cm}P{2.8cm}P{2.8cm}}
\raisebox{-0.4\height}{RGB} & 
\raisebox{-0.4\height}{Object Detection} & \raisebox{-0.4\height}{Semantic Segmentation} & \raisebox{-0.4\height}{Instance Segmentation} &
\raisebox{-0.4\height}{Panoptic Segmentation} &
\raisebox{-0.4\height}{Depth Estimation}  
\\
\raisebox{-0.4\height}{\includegraphics[width=\linewidth]{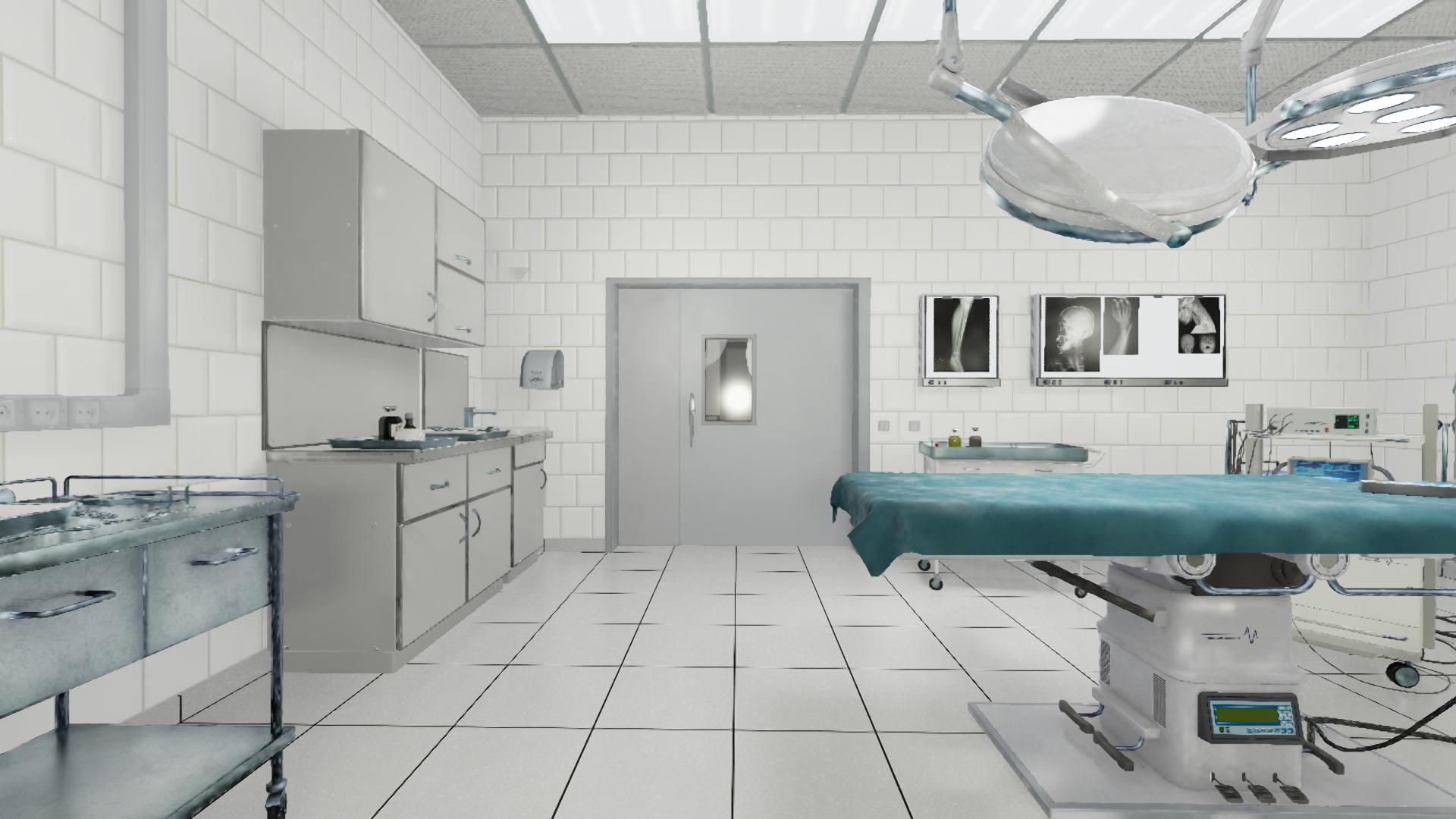}} & 
\raisebox{-0.4\height}{\includegraphics[width=\linewidth]{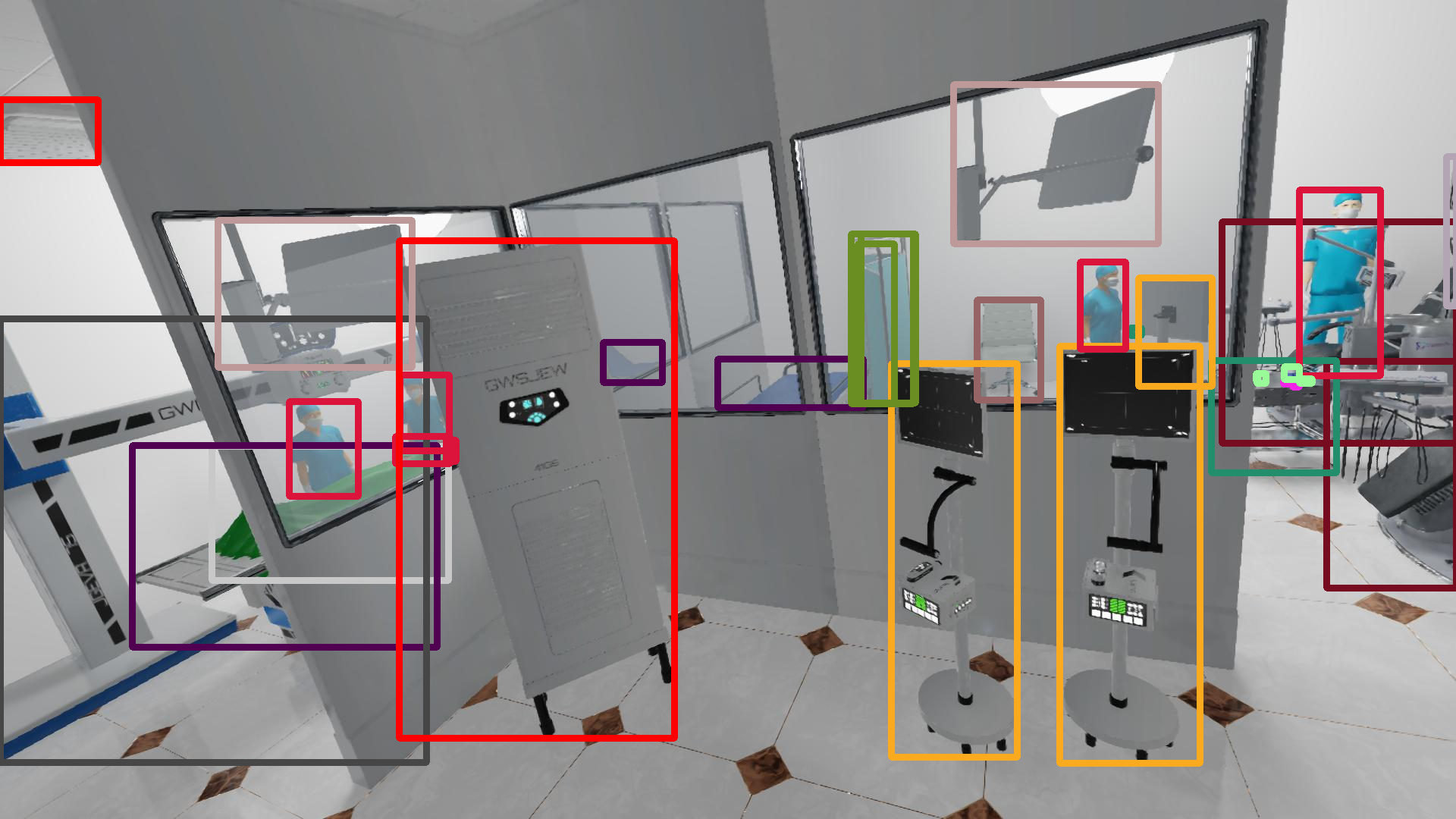}} & \raisebox{-0.4\height}{\includegraphics[width=\linewidth]{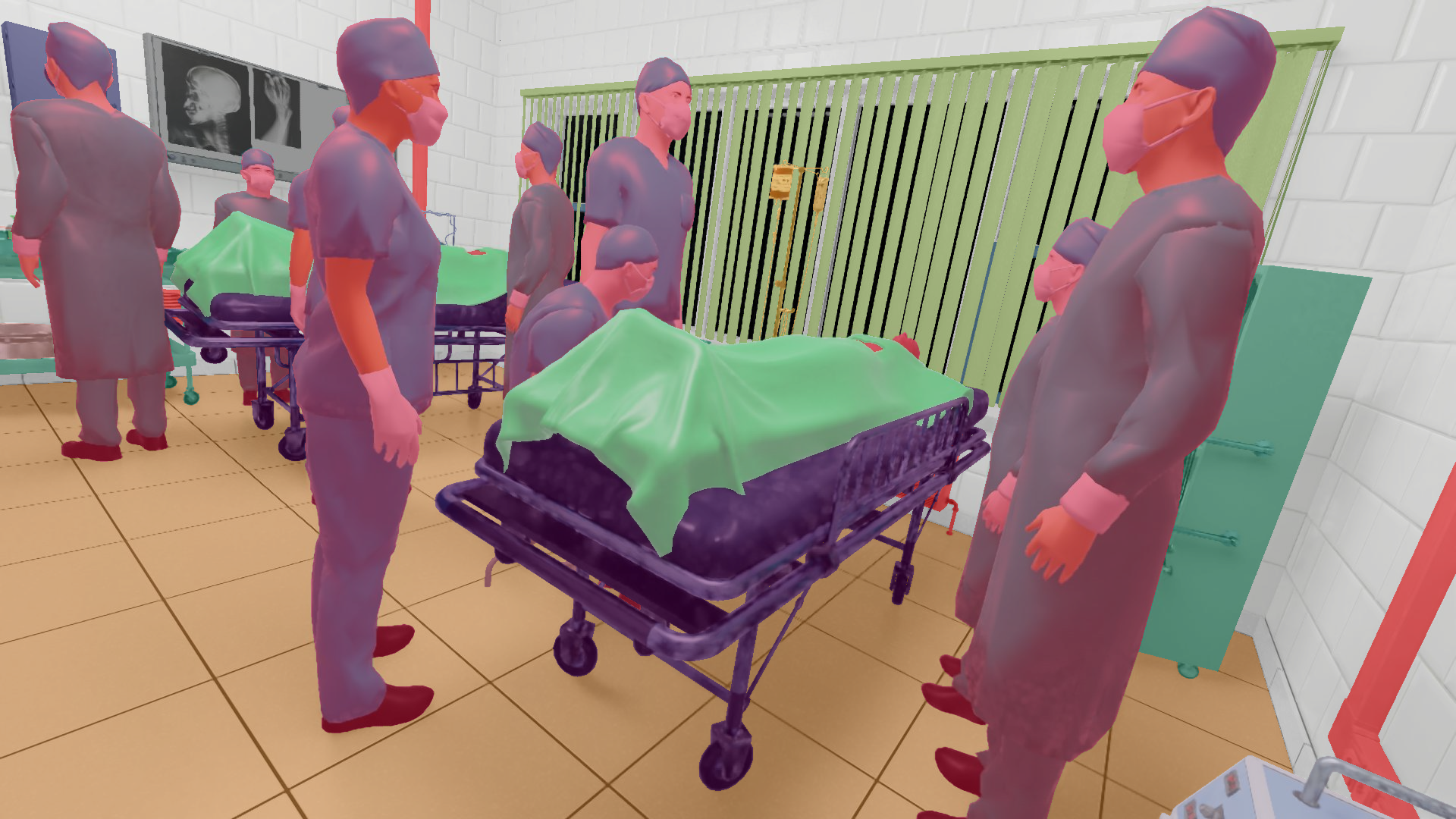}} & \raisebox{-0.4\height} {\includegraphics[width=\linewidth]{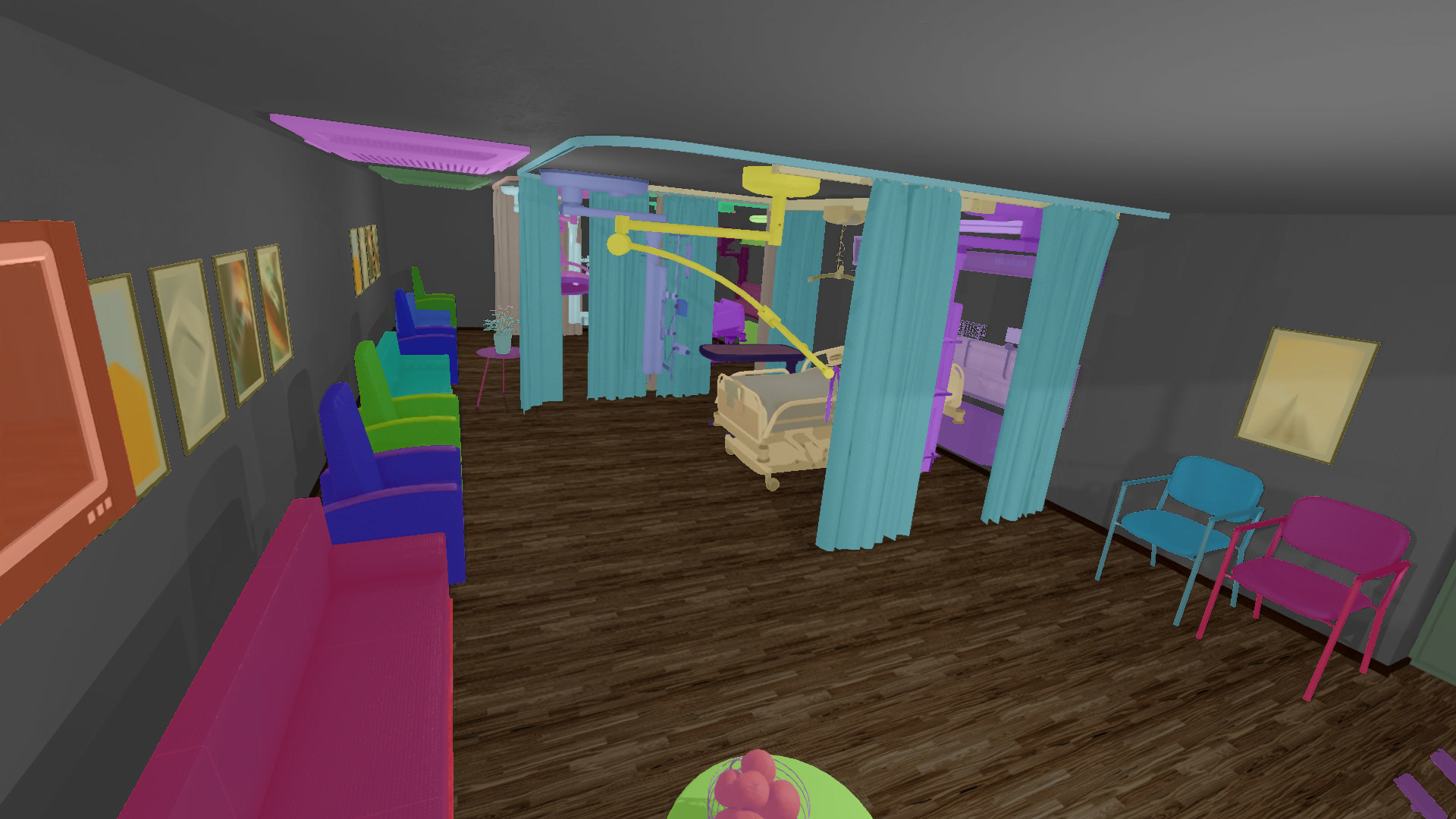}} &
\raisebox{-0.4\height}{\includegraphics[width=\linewidth]{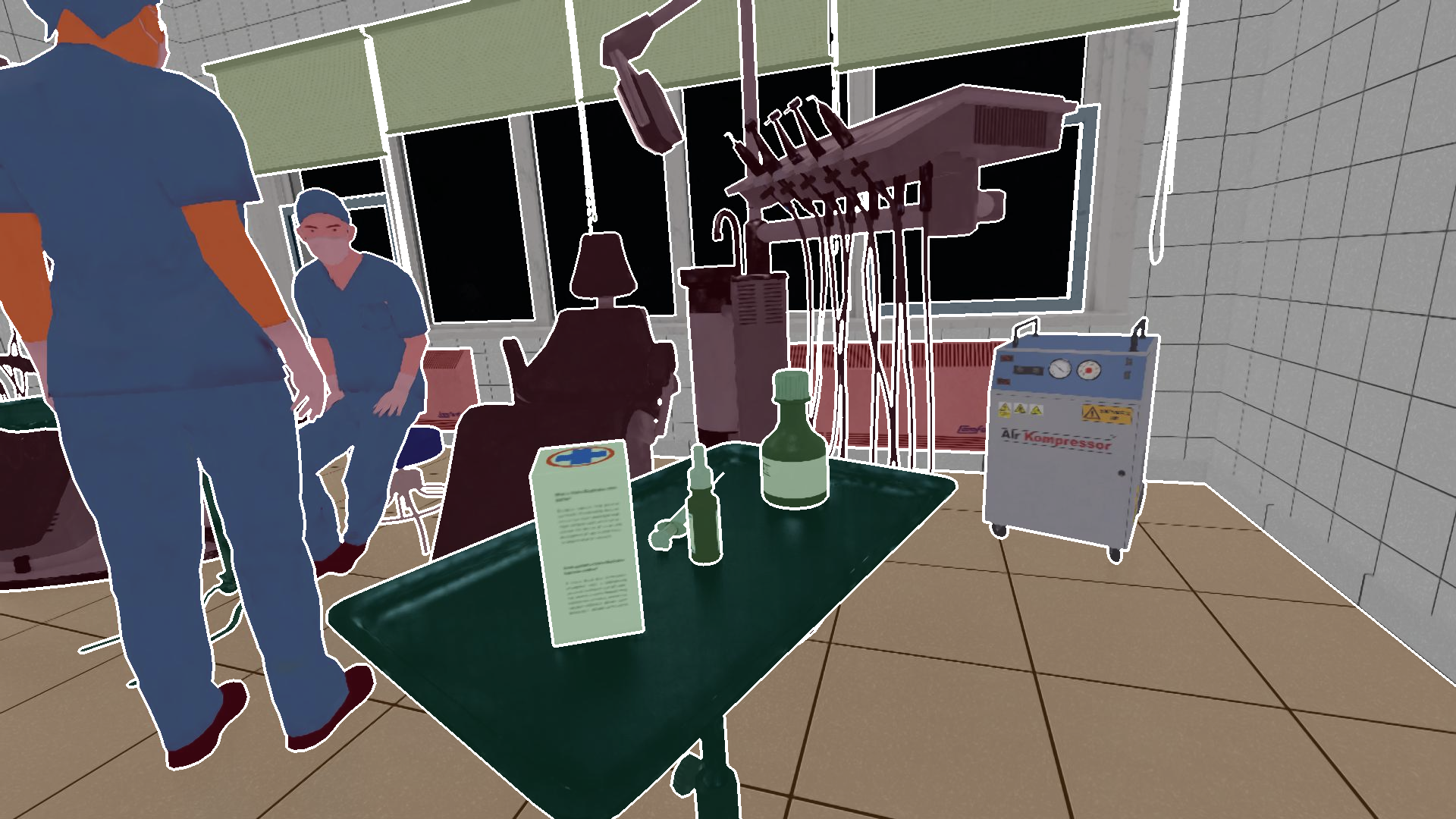}} &
\raisebox{-0.4\height}{\includegraphics[width=\linewidth]{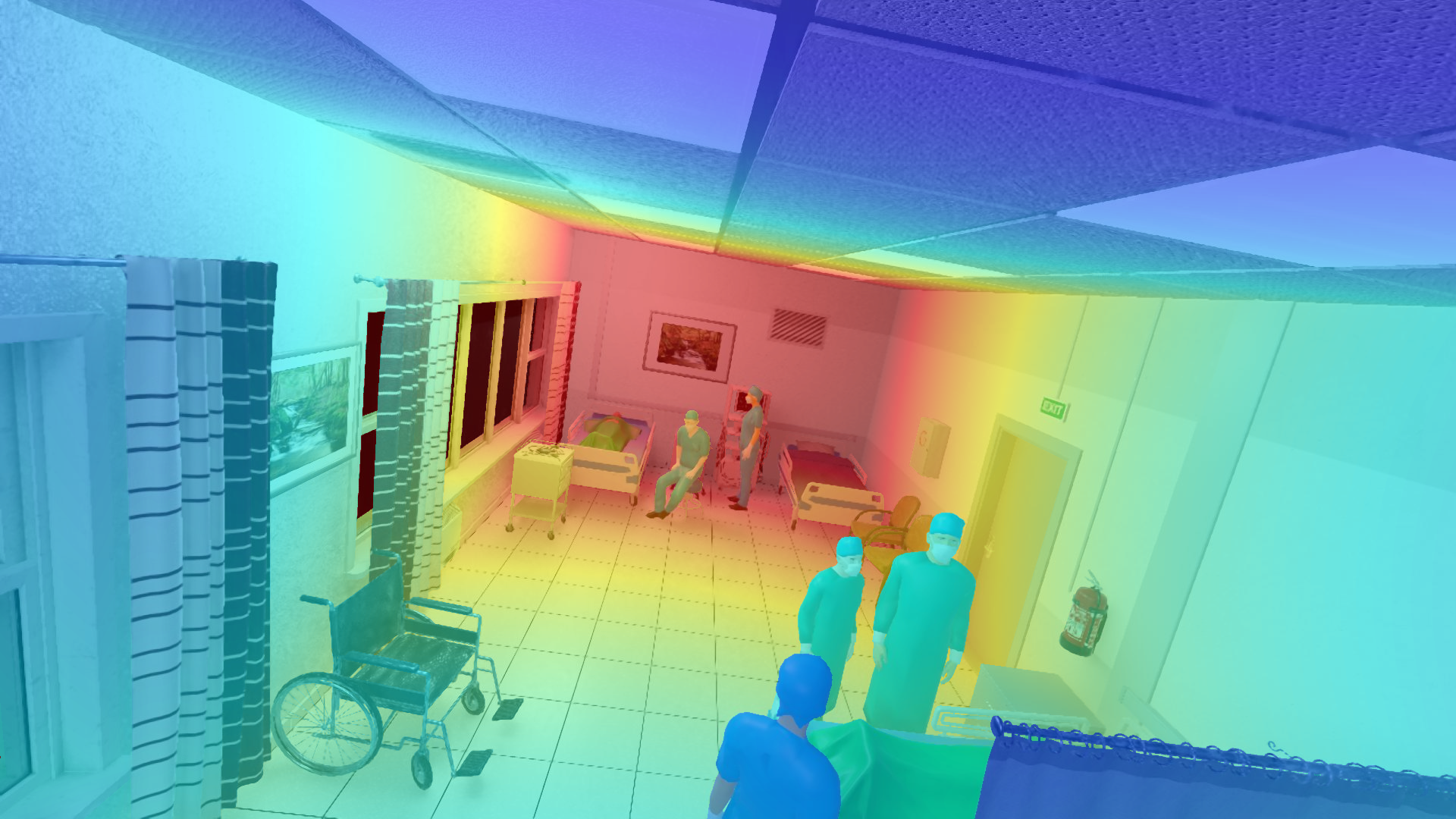}}  
\\
\\
\raisebox{-0.4\height}{\includegraphics[width=\linewidth]{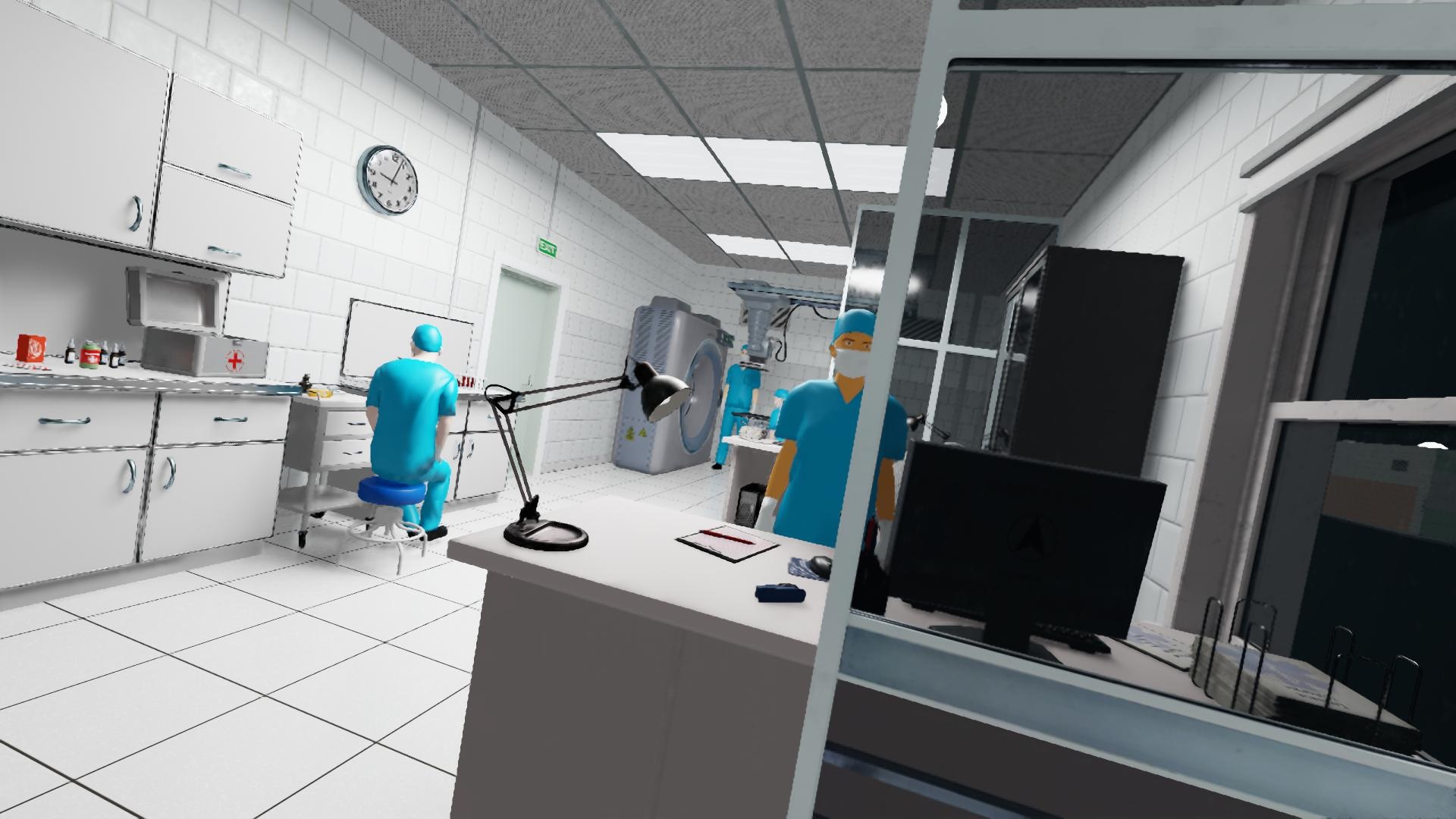}} & 
\raisebox{-0.4\height}{\includegraphics[width=\linewidth]{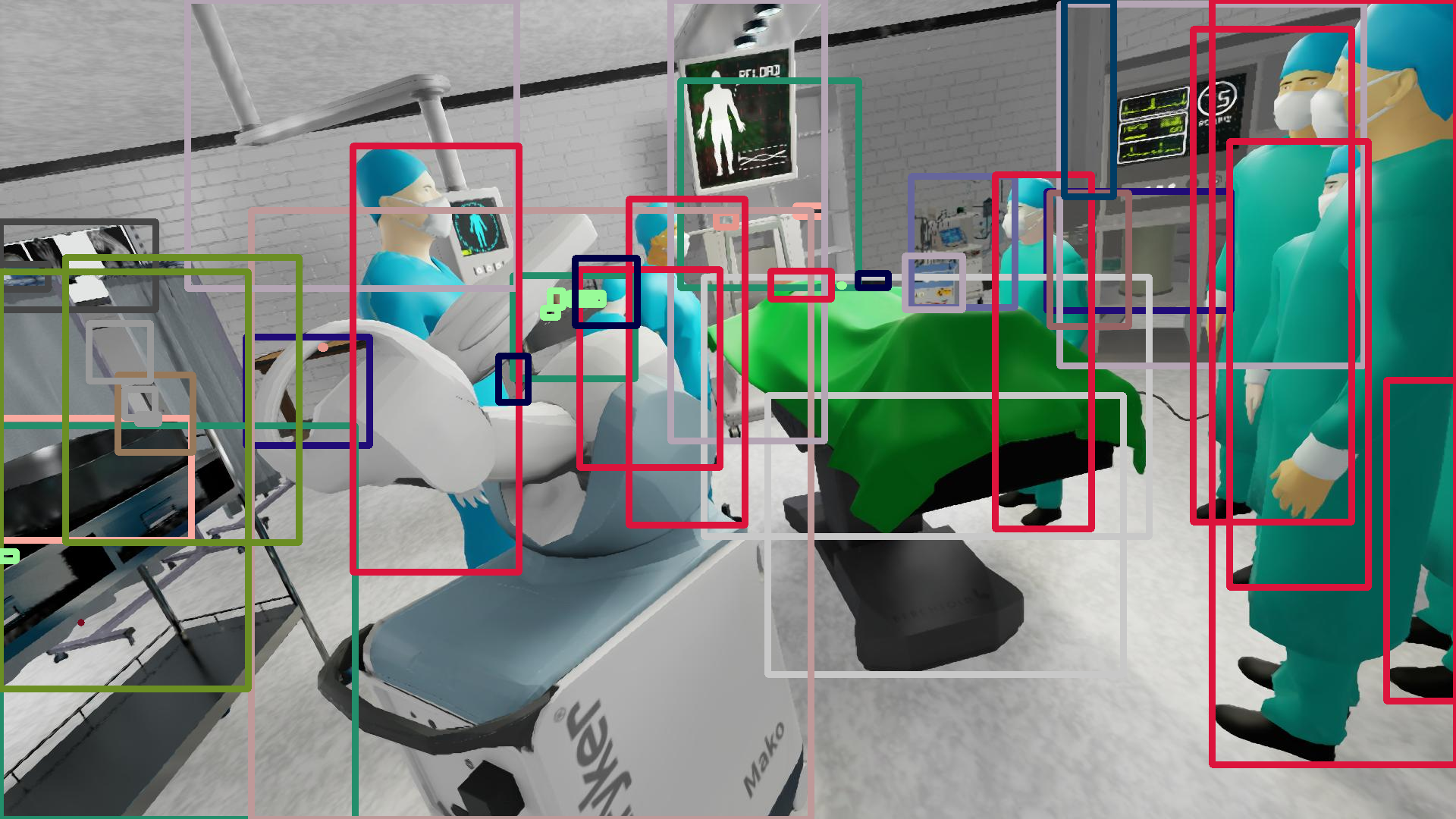}} & \raisebox{-0.4\height}{\includegraphics[width=\linewidth]{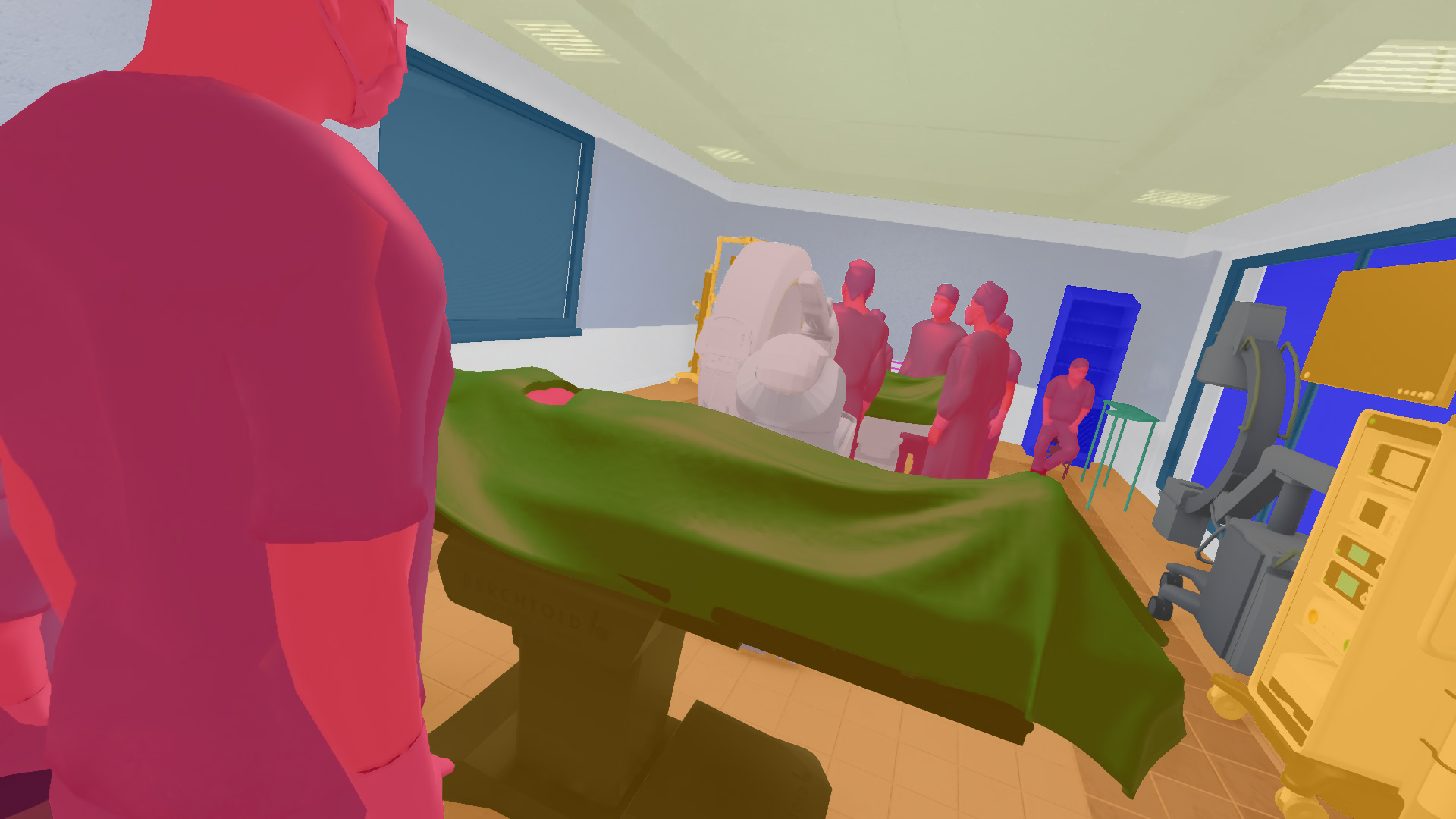}} & \raisebox{-0.4\height} {\includegraphics[width=\linewidth]{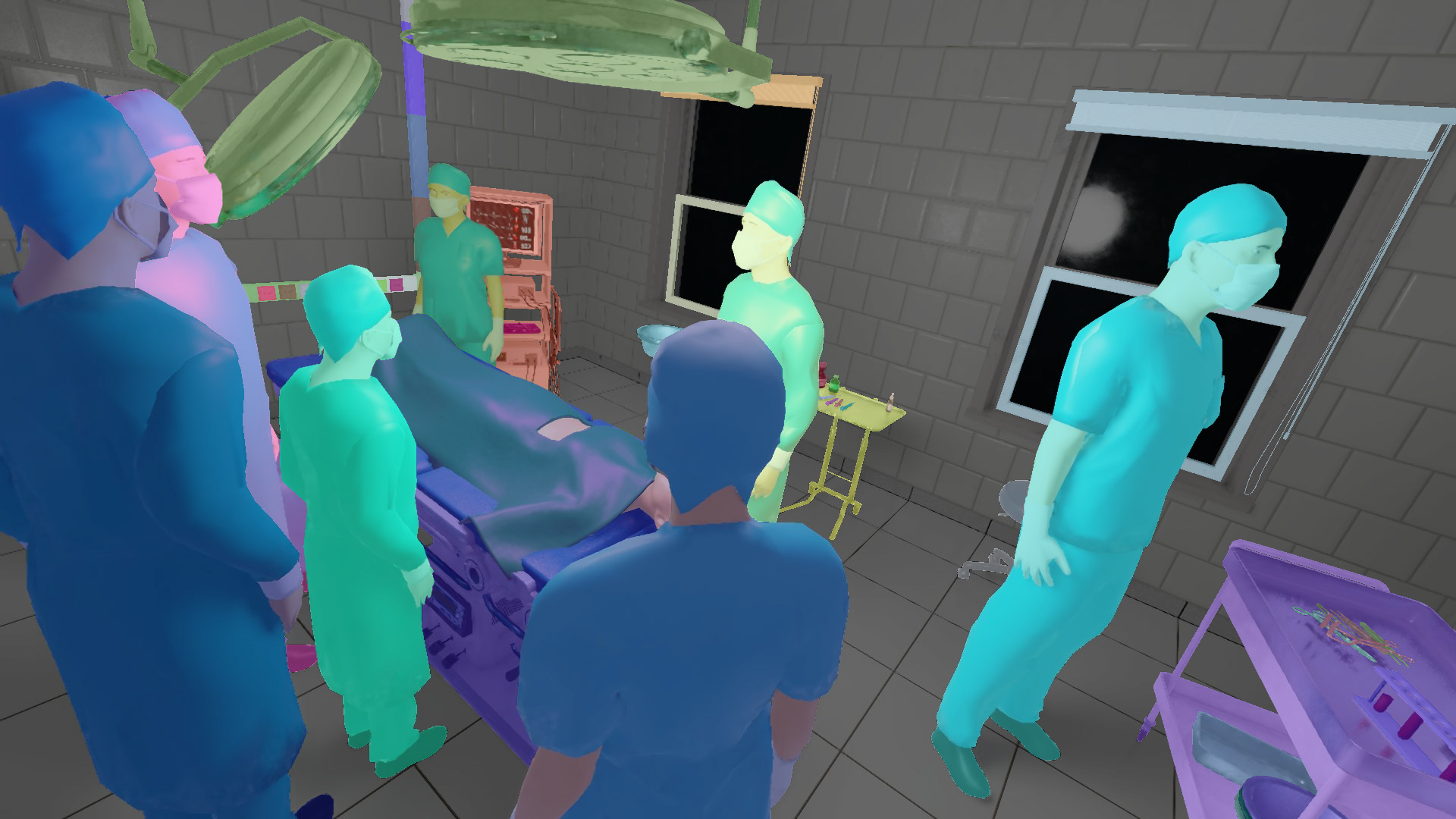}} &
\raisebox{-0.4\height}{\includegraphics[width=\linewidth]{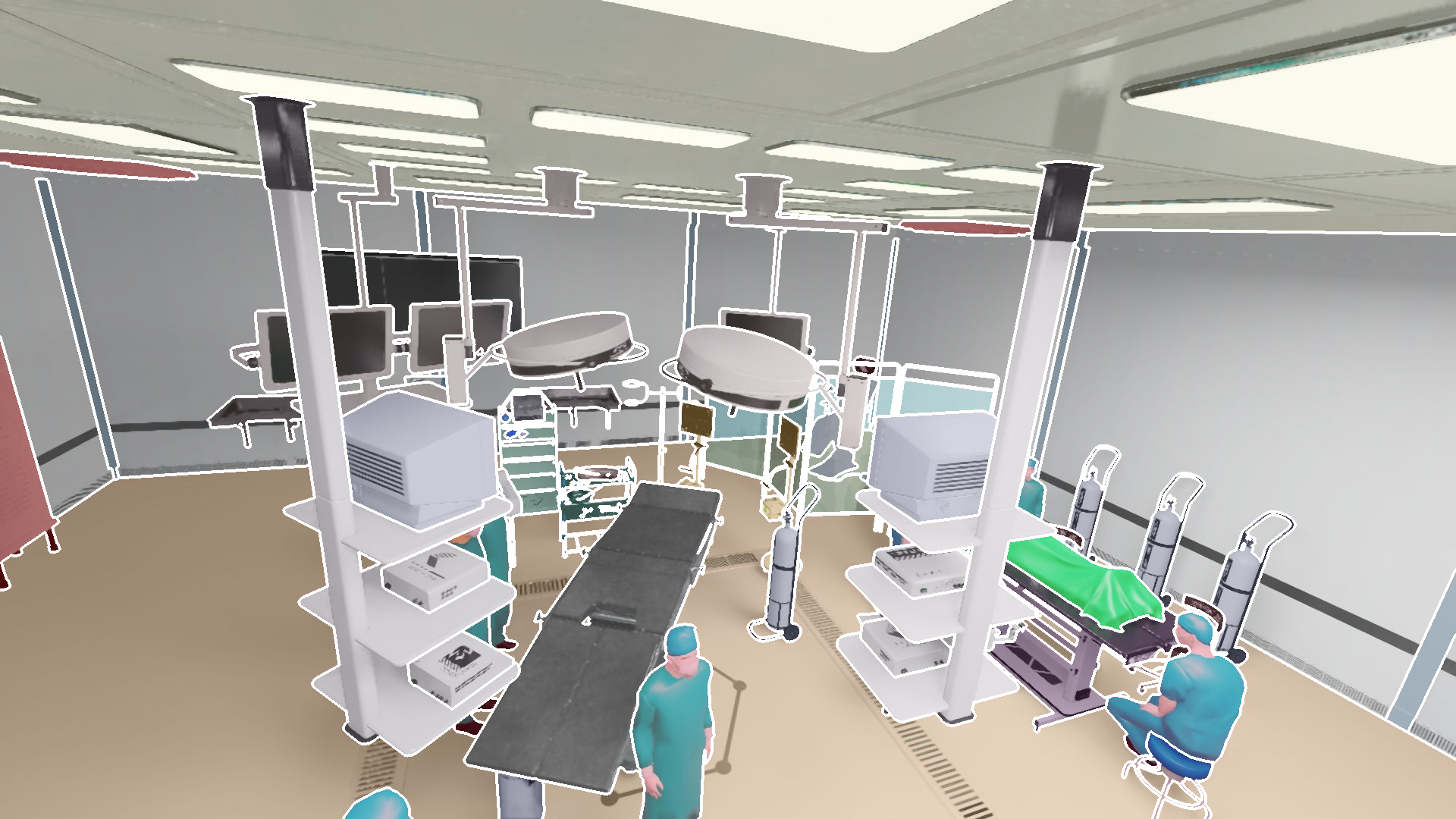}} &
\raisebox{-0.4\height}{\includegraphics[width=\linewidth]{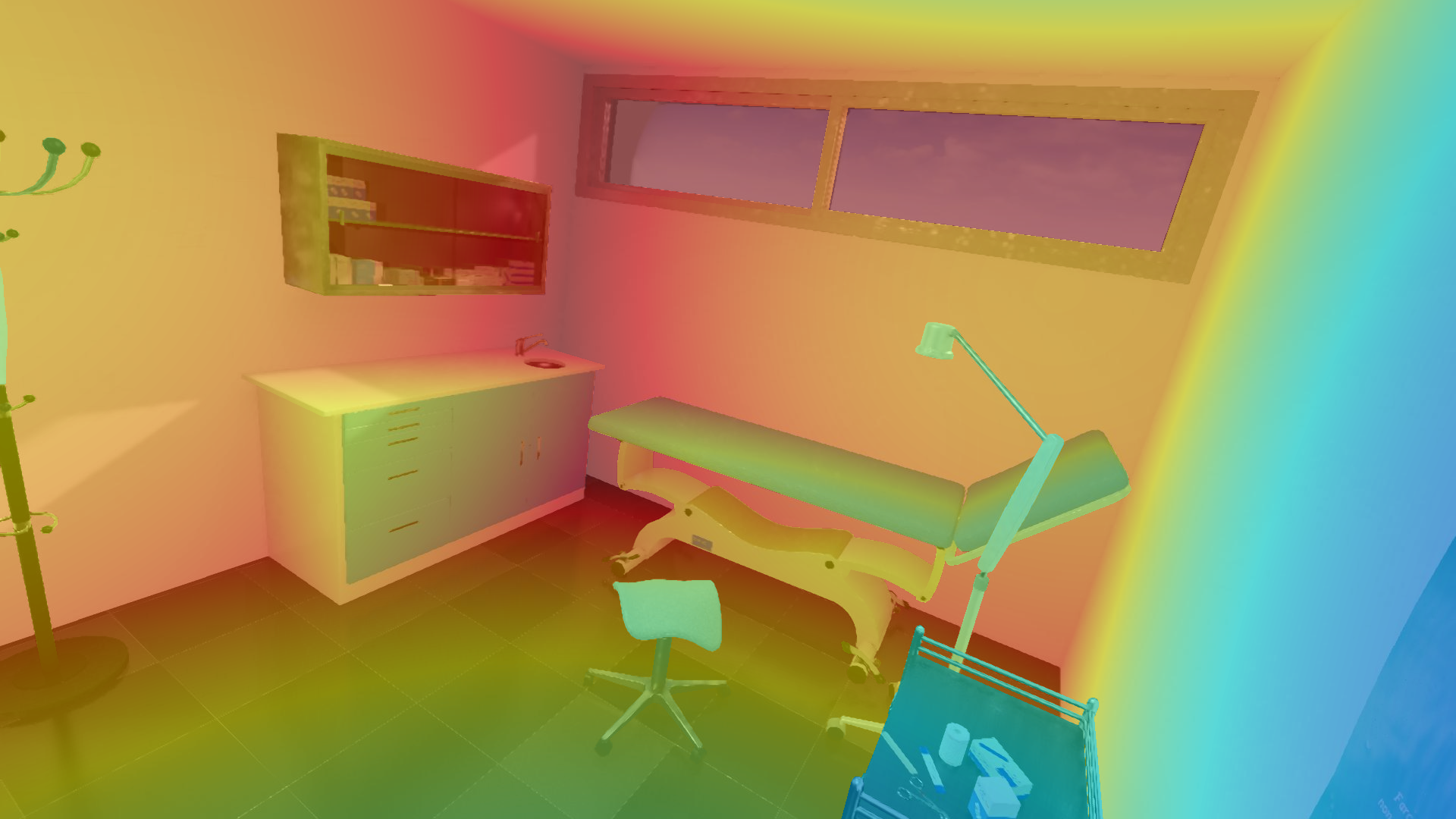}}  
\end{tabular}
\caption{Depiction of the diverse and complex environments in the \datasetname{} dataset. The images also show the ground truth labels for various tasks overlaid on the image. These images underline the comprehensive range of object types such as medical staff and equipment present in \datasetname{}, in addition to illustrating a variety of realistic scenarios through the interplay of lighting conditions and the interaction of elements within each scene. (8$\times$ zoom recommended.)}
\label{fig:datasetex}
\end{figure*}

\begin{table}
\footnotesize
\centering
\caption{List of semantic classes in the \datasetname{} dataset with examples of objects contained in the category. Please note that for all relevant benchmarking tasks, classes within the same row are unified as a single class.}
\label{tab:syn_class}
\footnotesize
\begin{tabular}{lll}
\toprule
&Semantic Class & Example \\
\toprule

\multirow[c]{35}{*}{\rotatebox[origin=c]{90}{\textit{Thing Classes}}} & Medical Chairs & surgical and dental chairs \\
& General Chairs & office and waiting chairs   \\
\cmidrule{2-3}
& Medical Tables & examination tables \\
\cmidrule{2-3}
&Medical Storage &  surgical shelves, trolleys \\
&General Storage & suitcase, box  \\
\cmidrule{2-3}
&Surgical Beds &  operating table   \\
&Medical Beds & stretcher, hospital bed \\
&Bed Accessories & cover, blanket \\ 
\cmidrule{2-3}
&Cutting Tools &  saw, knife \\ 
&Grasping Tools & clamp, tweezer  \\
&Support Tools & sterile container, tray   \\
&Measuring Tools & sphygmomanometer   \\
&Other Tools & picker, lancet  \\
\cmidrule{2-3}
&Advanced Assistance  &  robotic arm, airo system  \\ 
&Utility Equipment & medical pendant, air compressor  \\
\cmidrule{2-3}
&Imaging & x-ray scan  \\
\cmidrule{2-3}
&Life Support & defibrillator, infant incubator  \\
\cmidrule{2-3}
&Cabinet  & -  \\ 
&Cabinet Doors & -  \\
\cmidrule{2-3}
&Stand & clothes stand, camera stand  \\
\cmidrule{2-3}
&Person & doctor and patients  \\
\cmidrule{2-3}
&Curtain & -  \\
\cmidrule{2-3}
&Furniture \& Decor & painting, clipboard  \\
\cmidrule{2-3}
&Other Electronics & keyboard, mouse  \\
\cmidrule{2-3}
&Fixtures & fire alarm, info sign  \\
\cmidrule{2-3}
&Office Supplies & pen, stapler \\
\cmidrule{2-3}
&Building Elements & column, window, door \\
\midrule
\midrule
\multirow{5}{*}{\rotatebox[origin=c]{90}{\textit{Stuff Classes}}} &Wall & -  \\
\cmidrule{2-3}

&Floor & -  \\
\cmidrule{2-3}
&Ceiling & -  \\
\cmidrule{2-3}
&Miscellaneous & -  \\
\bottomrule
\end{tabular}
\end{table}

\subsection{Dataset Creation}
\label{subsec:creation}

The \datasetname{} dataset was generated using the NVIDIA Issac Sim simulator, a leading-edge simulation platform designed for robotics and AI research. This simulator provides direct access to ground truth labels for various computer vision tasks and presents relevant information about the simulated scene.
We captured our dataset using a multi-camera setup, modeled according to industry-standard cameras for surgical navigation such as NDI Polaris Vega~\cite{ndipolaris2023} and Stryker FP8000~\cite{stryker2023fp8000}.
The setup comprises three cameras---left, center, and right---arranged such that the distance between the left and right cameras is approximately \SI{422}{\centi\meter}. Both cameras are oriented to form an angle of \SI{9.5}{\degree} with the center camera. The precise intrinsic and extrinsic parameters of this camera setup are included with the dataset. \figref{FP8000_fig} illustrates the view from the multi-camera setup. The simulation environment comprises 13 distinct medical rooms including surgical suites, dental offices, radiology labs, and general hospital rooms. These environments are sourced from the Unreal Marketplace. During data collection, we bring in a variety of variations for each room. This includes diversity in interior surfaces, objects, lighting, and medical staff present in the scene, aiming to capture a wide spectrum of realistic scenarios. Moreover, to maintain consistency in the naming and semantic representation of assets across diverse environments, each type of asset is manually assigned a standardized name or semantic label. This approach ensures that identical semantics are uniformly applied across various environments. Examples of the rooms featured in our dataset are shown in \figref{fig:datasetex}. 

For data generation, we use two different methodologies. For the first approach, we designed over 20 unique camera trajectories per room by manually moving the cameras around the room in a human-like fashion. During the data collection phase, we set the initial position of the camera at varying locations and then randomly choose a trajectory from the existing pool to follow. The frames are captured at a rate of 10 FPS, which allows us to capture a variety of perspectives while minimizing redundancy. In the second methodology, we place the camera at various random positions without imposing any parametric constraints. This is followed by a manual curation process aimed at selecting the most representative data. This strategy broadens the diversity of data and enhances the robustness of models trained on this dataset. Lastly, we incorporate an additional layer of complexity by introducing noise through histogram matching using cumulative distribution functions. This procedure is specifically designed to adjust the color distribution of the synthetic images to match that of our private real-world data collected in ORs of a hospital. This additional noise serves to make the synthetic data more realistic and comparable to real-world data, improving its utility in terms of generalization and knowledge-transfer. \figref{init} shows an example of an initial image generated by the simulator, while \figref{with_noise} depicts the same image post noise addition.

\subsection{Dataset Structure}
\label{subsec:struct}

We captured a total of \num{16000} frames in the NVIDIA Isaac Sim simulation environment, each featuring three images---left, center, and right camera---with a resolution of $1920 \times 1080$ pixels, for a total of \num{48000} images. Our dataset includes 13 distinct medical rooms  which cater to a diverse range of scenarios. These scenarios account for varying illumination conditions, room configurations, and the presence of different medical equipment and staff.
We provide ground truth labels for 2D bounding boxes, semantic, instance, panoptic segmentation, and monocular depth estimation. \datasetname{} consists of 31 unique semantic classes, with instance-level annotations available for 27 of these classes. \tabref{tab:syn_class} outlines the semantic classes included in the dataset. 

The dataset is split strategically into training, validation, and test sets. The training set is composed of \num{9000} frames, while the validation and test sets contain \num{2000} and \num{5000} frames, respectively. This split is shared across all benchmarking tasks within the dataset. Importantly, to prevent hyperparameter fine-tuning and intentional overfitting, the annotations for the test set only cover the center camera, and will not be released publicly as they will be used to evaluate the performance of the models submitted to the online benchmark.
Therefore, we recommend using the validation subset for fine-tuning model parameters or selection before carrying out evaluations on the test server.

\subsection{Benchmarking}
\label{subsec:benchmark}

For benchmarking, we reduce the original 31 semantic classes of the dataset down to 21 classes. This is achieved by categorizing related classes into broader \textit{super} semantic classes. This strategic consolidation reduces the complexity of the task and aids in improving the performance and interpretability of the models. The adoption of more generalized classes provides a simpler framework for various task methodologies, allowing them to focus on larger categories rather than highly specific ones, which could potentially lead to overfitting. \datasetname{} provides a benchmark for six diverse tasks: object detection, semantic segmentation, instance segmentation, panoptic segmentation, as well as monocular depth estimation. This broad task range reflects the varying degrees of scene understanding required in different healthcare applications. Thus, whether an application requires bounding box object detection or detailed semantic segmentation, our dataset provides the necessary scope for development and testing. \figref{fig:datasetex} shows annotations overlaid on the camera image for each of the aforementioned benchmarking tasks. 

\begin{figure*}
    \centering
    \begin{subfigure}[b]{0.9\linewidth}
        \includegraphics[trim={150 0 400 0},clip,width=\linewidth]{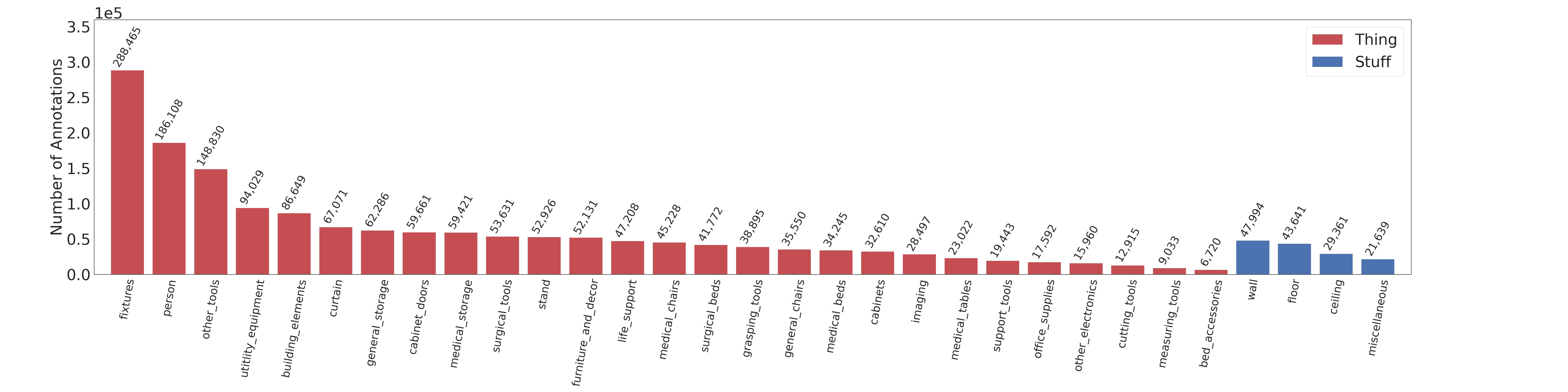}
        \subcaption{Number of segments per semantic class}
        \label{fig:inst_num}
    \end{subfigure}
        \begin{subfigure}[b]{0.3\linewidth}
        \includegraphics[width=\linewidth]{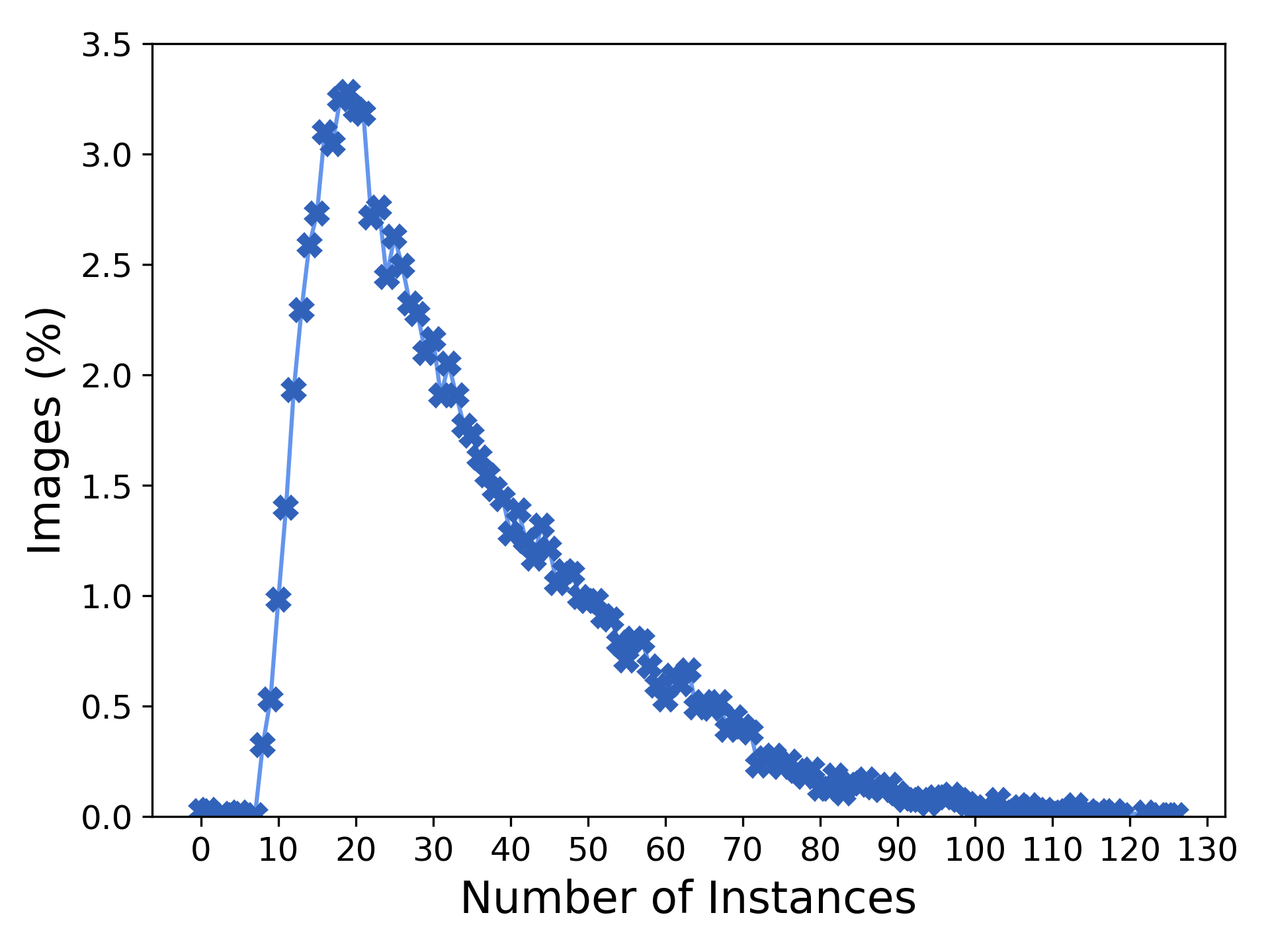}
        \subcaption{Distribution of images by instance count }
        \label{fig:occ_asd}
    \end{subfigure}
    \begin{subfigure}[b]{0.3\linewidth}
        \includegraphics[width=\linewidth]{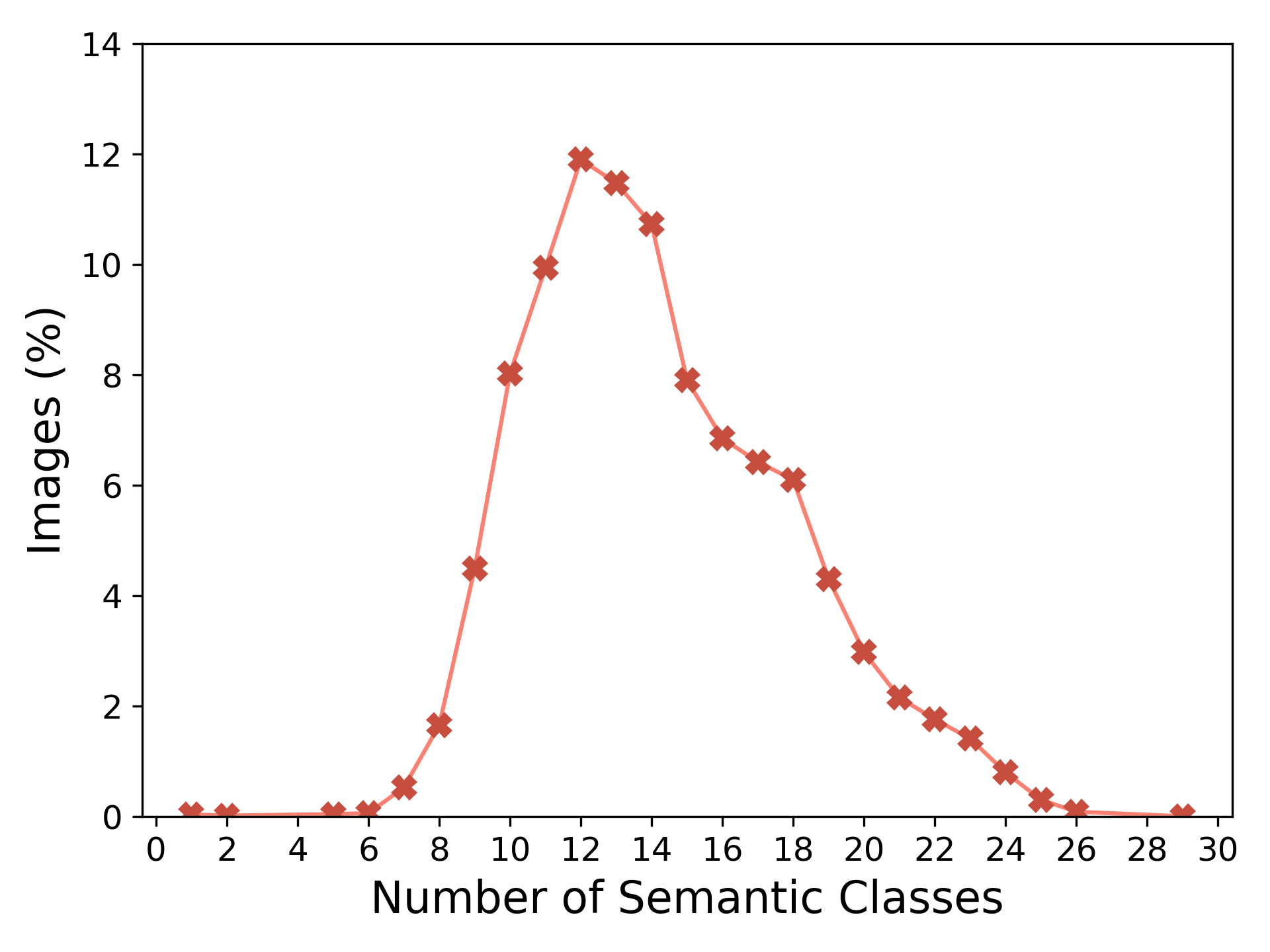}
        \subcaption{Distribution of images by semantic class}
        \label{fig:occ_asd1}
    \end{subfigure}
    \begin{subfigure}[b]{0.3\linewidth}
        \includegraphics[width=\linewidth]{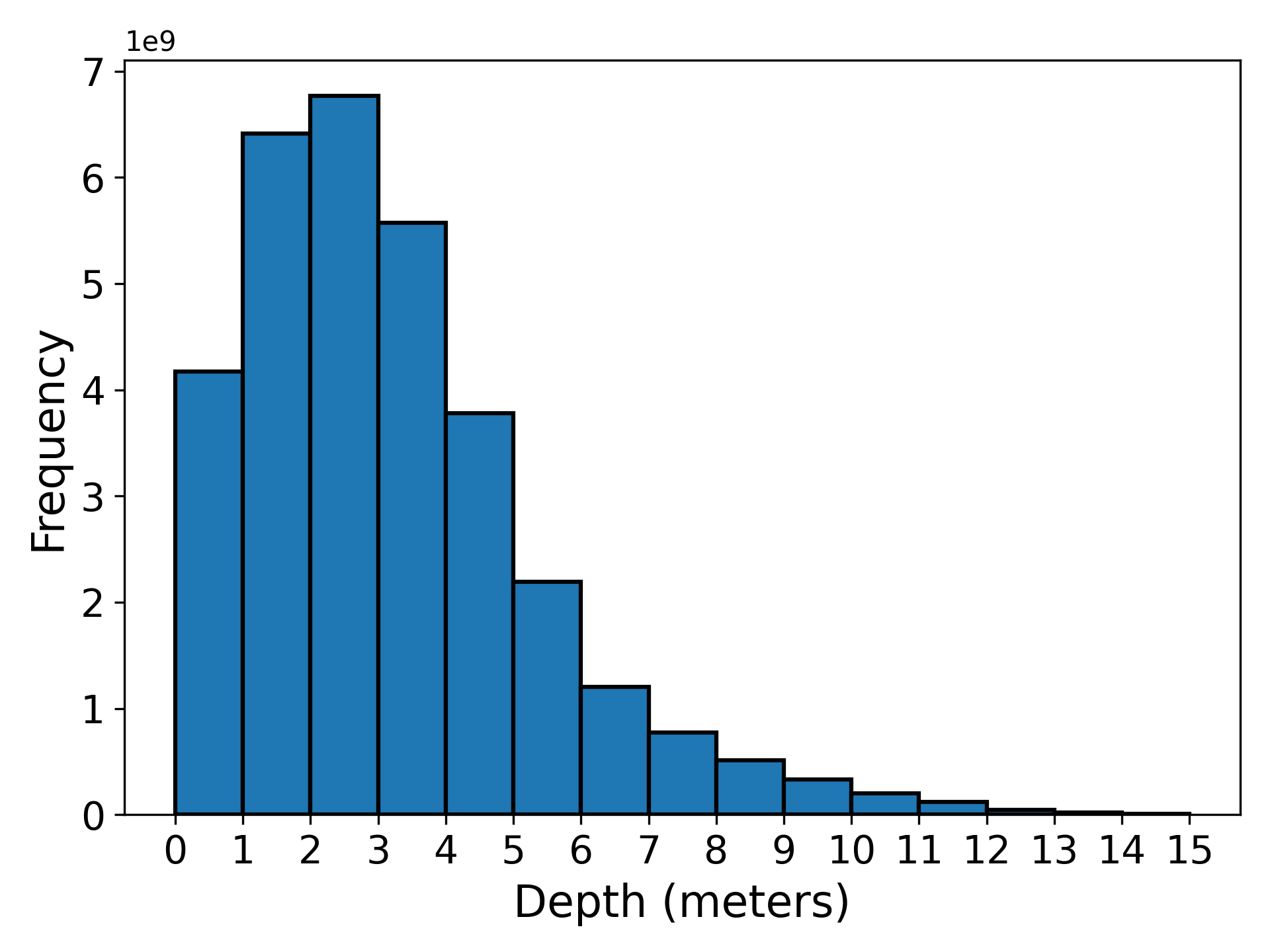}
        \subcaption{Depth-wise pixel distribution}
        \label{fig:occ_asd2}
    \end{subfigure}
    \caption{Statistical overview of the \datasetname{} dataset. Here, in (a) the distribution of object segments across semantic classes is shown, with blue bars for \textit{stuff} classes and red bars for \textit{thing} classes with instances. In (b) the distribution of images according to the instance count is depicted, reflecting the complexity of the scenes in the dataset. Following, (c) highlights the number of semantic classes present in each image, indicating the richness of class diversity. Lastly, (d) displays the depth-wise pixel distribution, underlining the depth variability in our dataset.}
    \label{fig:occ}
\end{figure*}

For object detection and instance segmentation, we conduct evaluations on 17 distinct semantic classes which fall under the \textit{thing} category as specified in \tabref{tab:syn_class}. These classes are chairs, medical tables, storage, beds, medical tools, medical equipment, imaging, life support, cabinet, stand, healthcare participants, curtain, furniture and decor, other electronics, fixtures, office supplies, and building elements. For semantic segmentation and panoptic segmentation, evaluations are performed across all 21 semantic classes including the 17~\textit{thing} classes and 4~\textit{stuff} classes: wall, floor, ceiling, and miscellaneous. Lastly, we evaluate monocular depth estimation up to a maximum depth of \SI{15}{\meter}. We set this limit due to a significant drop in pixel distribution near \SI{15}{\meter}, which is evident in \figref{fig:occ_asd2}. To quantify the performance in each task, we employ standard metrics as elaborated in the \secref{subsec:baseline_results}. We perform the benchmarking using a blend of both foundational models such as Faster-RCNN~\cite{girshick2015fast} and Mask R-CNN~\cite{he2017mask} as well as current state-of-the-art models such as Mask2Former~\cite{cheng2022masked} and BinsFormer~\cite{li2022binsformer}. The aim is to evaluate the versatility and effectiveness of our \datasetname{} dataset across these varying methodologies. The benchmarking results are intended to serve as a baseline for subsequent research with our introduced dataset.


\subsection{Data Statistics and Analysis}
\label{subsec:stats}

\figref{fig:occ} presents a detailed statistical overview of the dataset. \figref{fig:inst_num} depicts the distribution of object segments across the semantic classes; blue bars indicate \textit{stuff} classes and red bars represent \textit{thing} classes with distinct instances. Our dataset distinguishes itself with a diverse range of semantic classes, ranging from generic indoor objects to specific medical facilities. This richness and diversity in the dataset facilitate broad-spectrum scene understanding, enabling the development and testing of models intended for use in healthcare settings. The complexity of our dataset is further shown in \figref{fig:occ_asd} with the image distribution by instance count, where we see a significant peak between 10 to 50 instances. The maximum count of instances can even reach as high as 126. Furthermore,  \figref{fig:occ_asd1} presents the distribution of images relative to the semantic class count, peaking notably between 8 to 22 classes. This distribution suggests that each image typically captures a rich blend of semantic classes. Moreover, \figref{fig:occ_asd2} illustrates the depth-wise pixel distribution of our dataset, showing that most pixels associated with object segments are under \SI{5}{\meter}, with a gradual decrease observed as the depth increases towards the \SI{15}{\meter} mark. This depth distribution underscores the depth variation in our dataset. 

\section{Experimental Evaluation}
\label{sec:experiments}

In this section, we report the benchmark results for both the validation and test sets of Syn-Mediverse for all the tasks as outlined in~\secref{subsec:benchmark}. Subsequently, we present qualitative results on how semantics learned on our dataset can be transferred to real-world images. We perform all the experiments on images from a single viewpoint, specifically that of the center camera, which offers the most comprehensive view. We use PyTorch implementations of all the models and train on a system with two AMD EPYC 7452 (2.35GHz) processors and 8 NVIDIA GeForce RTX 3090 GPUs with a batch size of 16. Please refer to the supplementary material for more details on the training procedure of all the baselines.

\subsection{Baseline Results}
\label{subsec:baseline_results}

\begin{table}
\setlength\tabcolsep{3.0pt}
\centering
\caption{Comparison of object detection performance on the \datasetname{} dataset. All scores are in [\%].}
\label{tab:objectdet}
\footnotesize
\begin{tabular}
{l|ccc|ccc}
\toprule
\multirow{2}{*}{Method} & \multicolumn{3}{c|}{Val Set} & \multicolumn{3}{c}{Test Set} \\ 
\cmidrule{2-7}
 & mAP & mAP$_{50}$  & mAP$_{75}$ &mAP & mAP$_{50}$  & mAP$_{75}$ \\
\toprule
SSD~\cite{womg2018tiny} & 30.5& 55.5 & 30.0 &29.0 &53.0&28.4 \\
YOLO~\cite{redmon2018yolov3} & 43.4& 71.0& 46.0&41.2&68.0&43.6\\
Faster R-CNN~\cite{girshick2015fast} & 51.4 & 72.9 & 57.4&47.6&68.2&52.4 \\
DETR~\cite{zhu2020deformable} & 53.1 & 74.3 & 58.9&48.9&69.2&54.1\\
EfficientDet~\cite{tan2020efficientdet} & \textbf{55.6} & \textbf{78.9} & \textbf{61.2}& \textbf{53.7}&\textbf{75.2}&\textbf{59.1} \\

\bottomrule
\end{tabular}
\end{table}

\subsubsection{Object Detection}

We evaluate several object detection approaches on our \datasetname{} dataset, namely, SSD~\cite{womg2018tiny}, YOLO~\cite{redmon2018yolov3}, Faster R-CNN~\cite{girshick2015fast}, DETR~\cite{zhu2020deformable}, and EfficientDet~\cite{tan2020efficientdet}. We evaluate their performance in terms of mean average precision (mAP metric), mAP at an Intersection over Union (IoU) threshold of $0.50$ (mAP$_{50}$), and mAP at an IoU threshold of $0.75$ (mAP$_{75}$). \tabref{tab:objectdet} reports the results of this experiment. EfficientDet, a top-down approach, outperforms all the other models across all three evaluation metrics on both the validation and test sets. It achieves a mAP of $55.6\%$ on the validation set and $53.7\%$ on the test set. DETR achieves the next best mAP, followed by Faster R-CNN, YOLO, and SSD, respectively. The comparable performance on both the validation and test sets highlights our dataset's balanced complexity and diversity across splits which is free from any bias toward specific scenarios. The benchmarking results, while promising, also illustrates the potential for advancement, highlighting opportunities for future research in object detection within healthcare settings using our dataset.

\begin{table*}
\setlength\tabcolsep{3.0pt}
\begin{center}
\caption{Comparison of semantic segmentation performance on the \datasetname{} dataset. All scores are in [\%].} 
\label{tab:semseg}
\begin{tabular}{ll|ccccccccccccccccccccc|c}
\toprule
& Method & \begin{sideways}healthcare personel\end{sideways} & \begin{sideways}medical tables\end{sideways} & \begin{sideways}storage\end{sideways} & \begin{sideways}beds\end{sideways} & \begin{sideways}medical tools\end{sideways} & \begin{sideways}imaging\end{sideways} & \begin{sideways}life support\end{sideways} & \begin{sideways}medical equipment\end{sideways} & \begin{sideways}chairs\end{sideways} & \begin{sideways}cabinets\end{sideways} & \begin{sideways}stand\end{sideways} & \begin{sideways}curtain\end{sideways} & \begin{sideways}furniture \& decor\end{sideways} & \begin{sideways}other electronics\end{sideways} & \begin{sideways}fixtures\end{sideways} & \begin{sideways}office supplies\end{sideways}  & \begin{sideways}building elements\end{sideways} & \begin{sideways}wall\end{sideways}  &\begin{sideways}floor\end{sideways}  &\begin{sideways}ceiling\end{sideways}& \begin{sideways}misc.\end{sideways} & mIoU \\
\midrule
\multirow{5}{*}{\vertical{Val Set}} & DeepLabV3+ &  93.3 & 62.0&51.6& 74.8 & 38.5 & 68.9&48.9&61.6&52.6&56.1&52.1&83.8&81.4&49.0&78.7&67.8&64.9&85.3&86.9&88.2&85.2&68.2 \\
&HRNet~\cite{wang2020deep} &94.0&59.9&55.9&79.1&46.7&68.0&53.5&67.4&67.0&61.7&57.1&85.2&77.1&53.0&83.1&76.7&65.4&89.7&91.4&91.8&85.6&71.9 \\
&OCRNet~\cite{yuan1909segmentation} & 96.1&65.0&62.5&82.6&48.2&78.2&59.7&75.2&73.2&74.7&65.7&87.7&82.5&59.5&90.7&76.8&72.9&\textbf{92.7}&\textbf{92.7}&\textbf{94.2}&91.1&77.2  \\
&SETR~\cite{zheng2021rethinking} & 93.8&\textbf{68.5}&61.9&81.7&54.8&74.6&59.4&77.0&68.8&63.6&61.9&88.7&85.6&60.0&90.0&78.2&73.8&89.0&89.2&90.7&89.6&76.2  \\
&SegFormer~\cite{xie2021segformer} & \textbf{96.9}&67.5&\textbf{63.0}&\textbf{85.2}&\textbf{55.3}&\textbf{80.8}&\textbf{64.4}&\textbf{79.9}&\textbf{75.8}&\textbf{70.3}&\textbf{69.4}&\textbf{91.8}&\textbf{91.0}&\textbf{70.6}&\textbf{91.0}&\textbf{83.3}&\textbf{78.7}&91.2&89.8&90.9&\textbf{93.1}&\textbf{80.0}  \\
\midrule
\multirow{5}{*}{\vertical{Test Set}} &DeepLabV3+~\cite{chen2017rethinking} & 88.7 & 54.6&47.2&66.7&34.3&62.5&47.1&56.2&43.7&55.2&42.7&77.2&72.8&28.9&67.0&61.5&52.5&83.9&85.1&86.6&80.7&62.1 \\
&HRNet~\cite{wang2020deep} & 90.2&58.1&51.7&73.9&43.1&62.6&49.8&64.6&55.4&60.6&52.4&79.3&74.1&43.6&73.1&71.0&64.5&88.5&89.8&90.2&82.4&67.6 \\
&OCRNet~\cite{yuan1909segmentation} &92.6&62.8&57.4&77.5&45.7&72.9&56.7&69.9&65.4&66.5&58.4&83.3&76.4&52.9&84.2&75.6&70.7&\textbf{90.4}&\textbf{91.1}&\textbf{92.9}&86.3&72.8 \\
&SETR~\cite{zheng2021rethinking} & 92.4&65.5&58.6&78.9&48.0&71.8&56.0&74.0&67.4&\textbf{67.1}&58.4&88.1&84.6&\textbf{54.4}&\textbf{87.4}&75.9&72.1&88.8&88.6&88.9&85.2&74.0 \\
&SegFormer~\cite{xie2021segformer} & \textbf{94.7}&\textbf{66.8}&\textbf{63.7}&\textbf{79.4}&\textbf{51.3}&\textbf{76.3}&\textbf{63.4}&\textbf{75.3}&\textbf{73.6}&65.9&\textbf{63.5}&\textbf{90.6}&\textbf{88.6}&51.9&87.2&\textbf{79.7}&\textbf{76.7}&89.9&87.1&91.1&\textbf{87.3}&\textbf{76.4}\\
\bottomrule
\end{tabular}
\end{center}
\end{table*}

\subsubsection{Semantic Segmentation}

\tabref{tab:semseg} presents the performance comparison of five widely adopted semantic segmentation models on the \datasetname{} dataset. We use the mean Intersection-over-Union (mIoU) for the evaluations, which is the standard metric for evaluating semantic segmentation performance. The results readily indicate that the benchmarked models handle certain classes more effectively than others, underlining the diversity and complexity of scenarios present in our dataset. From the metric scores of the individual categories, we observe that high-level amorphous regions such as floors, ceilings, and walls tend to have higher mIoU scores across all the methods, as these elements are relatively consistent in their visual properties. In contrast, more complex categories such as medical tools, life support, and storage tend to score lower due to their varying appearances and complex shapes. SegFormer~\cite{xie2021segformer} demonstrates the most robust performance across most categories, achieving a mIoU score of $80.0\%$ on the validation set and $76.4\%$ on the test set, thus attesting to its effectiveness in handling a diverse range of categories under different conditions. Interestingly, although all models show a decline in performance from validation to testing, the consistency in this performance drop across classes and models once again corroborates the balanced nature of our dataset.

\begin{table}
\setlength\tabcolsep{3.0pt}
\centering
\caption{Comparison of instance segmentation performance on the \datasetname{} dataset. All scores are in [\%].}
\label{tab:instanceseg}
\footnotesize
\begin{tabular}
{l|ccc|ccc}
\toprule
\multirow{2}{*}{Method} & \multicolumn{3}{c|}{Val Set} & \multicolumn{3}{c}{Test Set} \\ 
\cmidrule{2-7}
 & mAP & mAP$_{50}$  & mAP$_{75}$ &mAP & mAP$_{50}$  & mAP$_{75}$ \\
\toprule
 YOLOACT~\cite{bolya2019yolact} & 33.0& 55.8& 33.8&29.3&50.1&29.5\\
 Mask R-CNN~\cite{he2017mask} & 36.6& 59.8& 37.9&32.8&54.2&33.5  \\
 SOLOv2~\cite{wang2020solov2} & 37.0& 60.5& 37.7&37.2&59.3&38.1\\
DetectorRS~\cite{qiao2021detectors} & \textbf{44.1} & \textbf{66.3} & \textbf{44.9}&\textbf{42.4}&\textbf{63.3}&\textbf{44.1}\\
\bottomrule
\end{tabular}
\end{table}

\subsubsection{Instance Segmentation}

We evaluate the efficacy of our dataset for instance segmentation by benchmarking four models: YOLOACT~\cite{bolya2019yolact}, Mask R-CNN~\cite{he2017mask}, SOLOv2~\cite{wang2020solov2}, and DetectorRS~\cite{qiao2021detectors}. Similar to object detection, the key performance metric is the mean average precision (mAP), with additional insights provided at IoU thresholds of $0.50$ (mAP$_{50}$) and $0.75$ (mAP$_{75}$), as outlined in \tabref{tab:instanceseg}. For this task, DetectorRS proves to be the most effective approach, outperforming other models in all evaluation aspects on both validation and testing sets. It achieves a mAP of $44.1\%$ on the validation set and $42.4\%$ on the testing set, suggesting that our dataset is complex and challenging, even for high-performing models. We also observe that all the models other than SOLOv2 suffer a performance drop when moving from val to test. This highlights the robustness of SOLOv2 as well as our dataset's capability to identify models not based on only performance but generalizability in the scope of healthcare environments.
 
\begin{table}
\setlength\tabcolsep{1.8pt}
\centering
\caption{Comparison of panoptic segmentation performance on the \datasetname{} dataset. All scores are in [\%].}
\label{tab:Panoptic}
\footnotesize
\begin{tabular}
{l|ccccc|ccccc}
\toprule
\multirow{3}{*}{Method} & \multicolumn{5}{c|}{Val Set} & \multicolumn{5}{c}{Test Set} \\ 
\cmidrule{2-11}
& PQ & SQ  & RQ & PQ\textsuperscript{Th} & PQ\textsuperscript{St}& PQ & SQ  & RQ & PQ\textsuperscript{Th} & PQ\textsuperscript{St}\\
\toprule
Seamless~\cite{porzi2019seamless} & 60.0 & 85.4&  69.3 & 54.0 & 85.5 & 58.2 &84.8&67.6&52.0& 84.4 \\
P-DeepLab~\cite{cheng2020panoptic} & 62.2 & 85.8 & 71.6 & 56.6 & 85.9 & 60.3 & 85.2 & 70.0 & 54.6 & 84.7 \\
EfficientPS~\cite{mohan2021efficientps} &63.9 & 86.0& 73.5 & 58.7 & 86.1 & 61.6 & 85.3 & 71.4 &56.2 & 84.7 \\
Mask2Former~\cite{cheng2022masked} & \textbf{67.1}&\textbf{86.6}& \textbf{76.7}&\textbf{62.2} &\textbf{87.7} & \textbf{65.0}&8\textbf{6.1}&\textbf{74.8}&\textbf{60.1}&\textbf{86.2} \\
\bottomrule
\end{tabular}
\end{table}

\subsubsection{Panoptic Segmentation}

For panoptic segmentation on the \datasetname{} dataset, we compare four 
methods: Seamless~\cite{porzi2019seamless}, Panoptic-DeepLab~\cite{cheng2020panoptic}, EfficientPS~\cite{mohan2021efficientps}, and Mask2Former~\cite{cheng2022masked}. The results from this experiment are presented in \tabref{tab:Panoptic}. We primarily quantify the performance using the Panoptic Quality(PQ) metric. We also report Segmentation Quality (SQ), Recognition Quality (RQ), and the PQ equivalents for \textit{thing} (PQ\textsuperscript{Th}) and \textit{stuff} (PQ\textsuperscript{St}) for additional insights. \tabref{tab:Panoptic} presents the results for this experiment. We observe that Mask2Former outperforms all the other baselines by achieving a PQ score of 67.1\% on the validation set and 65.0\% on the test set. Thus, demonstrating the superiority of the recent transformer-based models in understanding complex scenes. From the PQ\textsuperscript{St} scores, we infer that the models are generally better at detecting \textit{stuff} classes, which are typically more uniform and have continuous areas. While the lower PQ\textsuperscript{Th} scores indicate that \datasetname{} poses a significant challenge when it comes to fine-grained object recognition and segmentation.

\subsubsection{Monocular Depth Estimation}

We evaluate the performance of three monocular depth estimation methods as baselines: BinsFormer~\cite{li2022binsformer}, DepthFormer~\cite{li2022depthformer}, and SimIPU~\cite{li2022simipu}. Given that the frames from all three cameras have depth ground truth, we train the models by treating them as independent. In \tabref{tab:monoDepth}, we present the results with the following metrics: Absolute Relative error (AbsRel), Root Mean Square Error (RMSE), Scale Invariant Logarithmic Error (SILog), and the percentage of inlier pixels within a 25\% ($\delta_1$), 56.25\% ($\delta_2$), and 95.31\% ($\delta_3$) depth estimation error. We observe from the results that DepthFormer performs the best across all the metrics for both data splits. While both BinsFormer~\cite{li2022binsformer} and SimIPU~\cite{li2022simipu} show good performance on the validation set, their performance significantly deteriorates on the test set. This suggests potential overfitting to the specific characteristics of the validation set or a lack of generalizability to unseen data.

\newcommand{\bst}[1]{\textbf{#1}}
\begin{table}
\setlength{\tabcolsep}{3pt}
\centering
\caption{Comparison of monocular depth estimation performance on the \datasetname{} dataset.}
\label{tab:monoDepth}
\footnotesize
\begin{tabular}
{ll|ccc|ccccccccc}
\toprule
& \thead{Method} & \thead{AbsRel \\ $[\%]\,\downarrow$} & \thead{RMSE \\ $[m]\,\downarrow$}  & \thead{SILog \\ $[\%]\,\downarrow$} & \thead{$\delta_1$ \\ $[\%]\,\uparrow$} & \thead{$\delta_2$ \\ $[\%]\,\uparrow$} & \thead{$\delta_3$ \\ $[\%]\,\uparrow$} \\
\midrule
\multirow{3}{*}{\vertical{Val Set}} 
& BinsFormer~\cite{li2022binsformer}   &      62.9  &      0.573  &      32.5  &      58.3  &      84.3  &      92.7 \\
& DepthFormer~\cite{li2022depthformer} & \bst{26.5} & \bst{0.469} & \bst{23.6} & \bst{69.1} & \bst{91.9} & \bst{96.9} \\
& SimIPU~\cite{li2022simipu}           &      55.2  &      0.693  &      38.3  &      51.9  &      80.4  &      90.5 \\
\midrule
\multirow{3}{*}{\vertical{Test Set}}
& BinsFormer~\cite{li2022binsformer}   &      60.9  &      1.740  &      44.4  &      32.1  &      56.3  &      72.7  \\
& DepthFormer~\cite{li2022depthformer} & \bst{39.4} & \bst{1.390} & \bst{41.3} & \bst{42.6} & \bst{65.9} & \bst{79.3} \\
& SimIPU~\cite{li2022simipu}           &      66.5  &      1.899  &      58.3  &      29.4  &      50.0  &      64.7  \\
\bottomrule
\end{tabular}
\end{table}




\begin{figure}
\centering
\footnotesize
\setlength{\tabcolsep}{0.1cm}
\begin{tabular}{P{0.5cm}P{3.5cm}P{3.5cm}}
& 
\raisebox{-0.4\height}{Input} & \raisebox{-0.4\height}{Semantic Prediction} 
\\
\vertical{(a) MVOR} & 
\raisebox{-0.4\height}{\includegraphics[width=\linewidth]{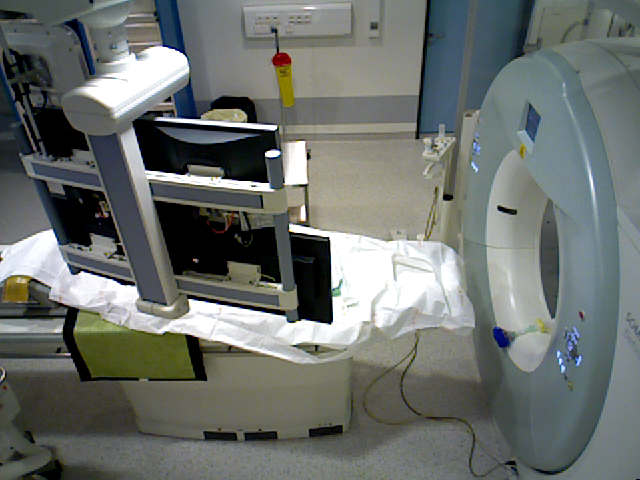}} & \raisebox{-0.4\height}{\includegraphics[width=\linewidth]{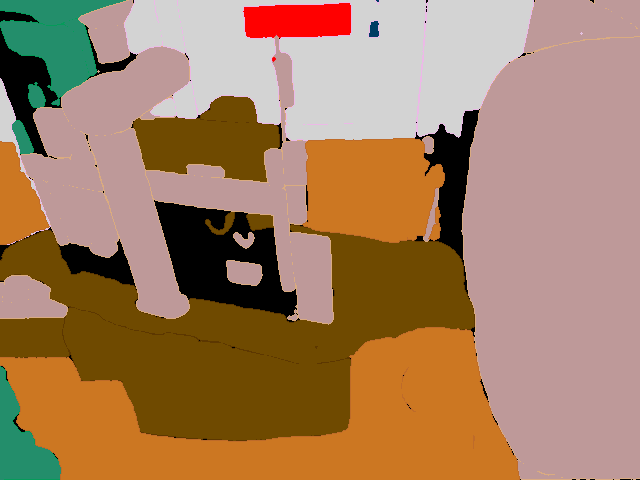}} 
\\
\\
\vertical{(b) MVOR} & 
\raisebox{-0.4\height}{\includegraphics[width=\linewidth]{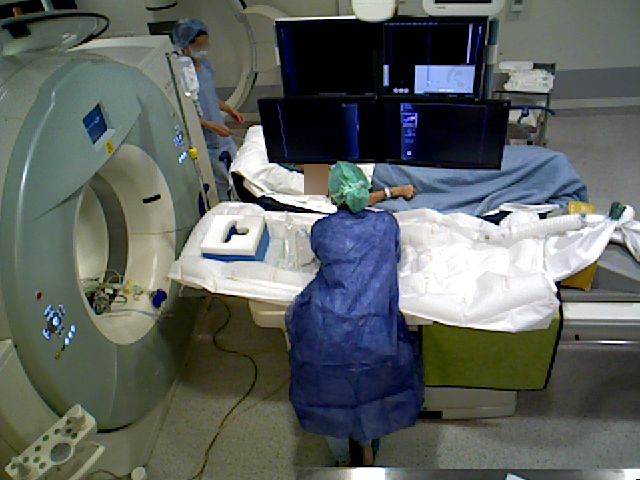}} & \raisebox{-0.4\height}{\includegraphics[width=\linewidth]{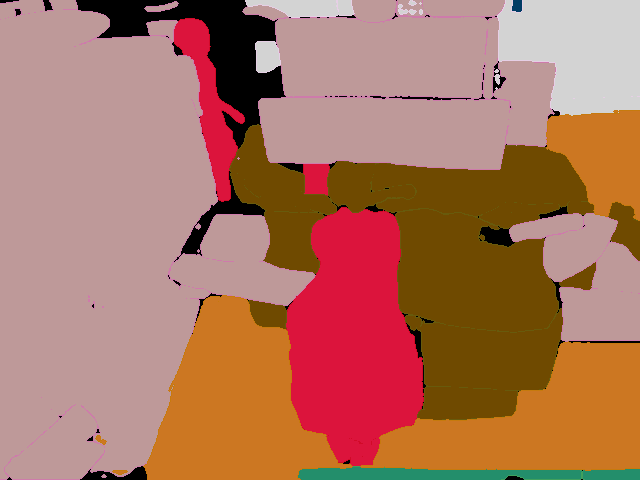}} \\
\\
\vertical{(c) 4D-OR} & 
\raisebox{-0.4\height}{\includegraphics[width=\linewidth]{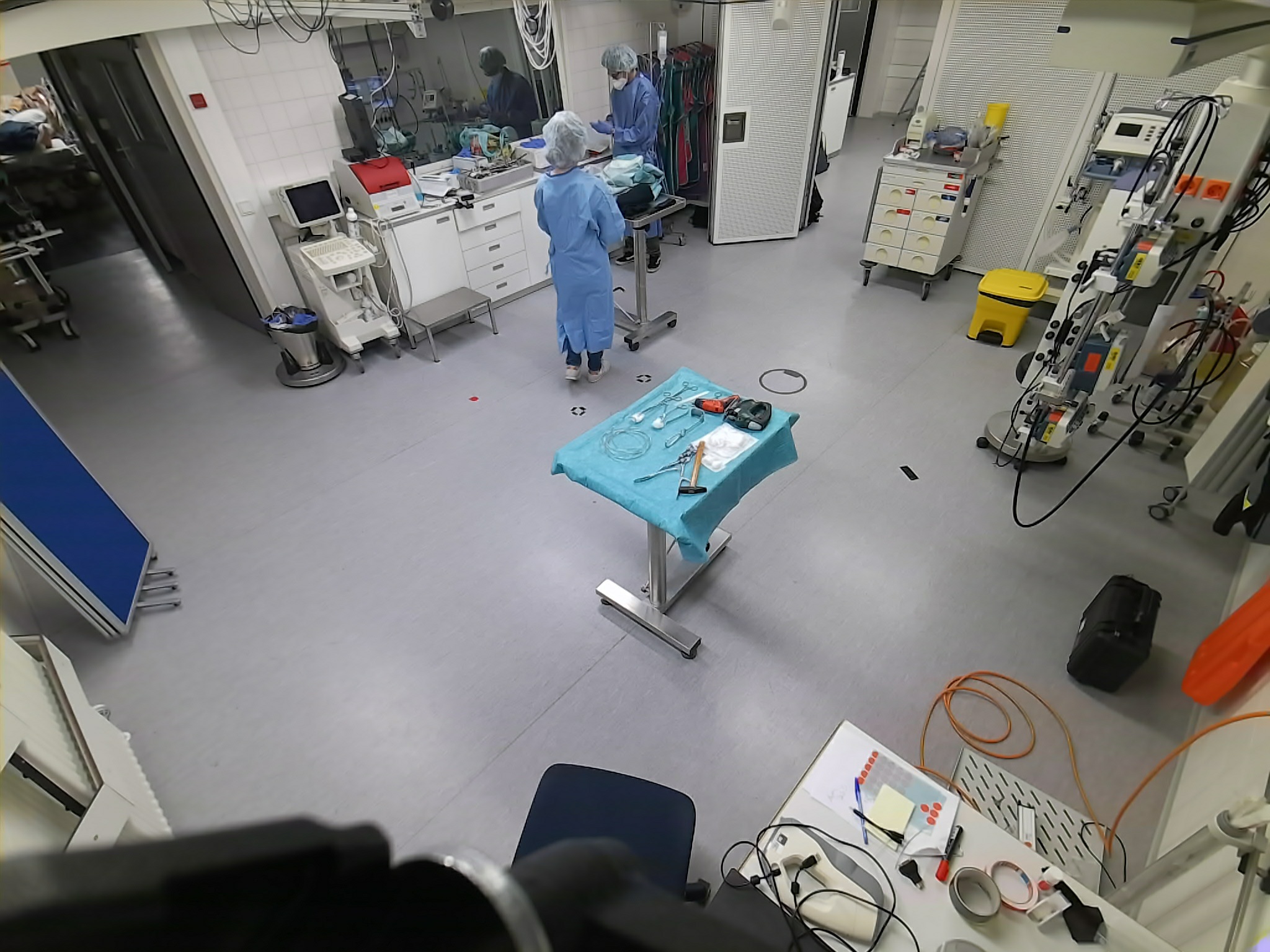}} & \raisebox{-0.4\height}{\includegraphics[width=\linewidth]{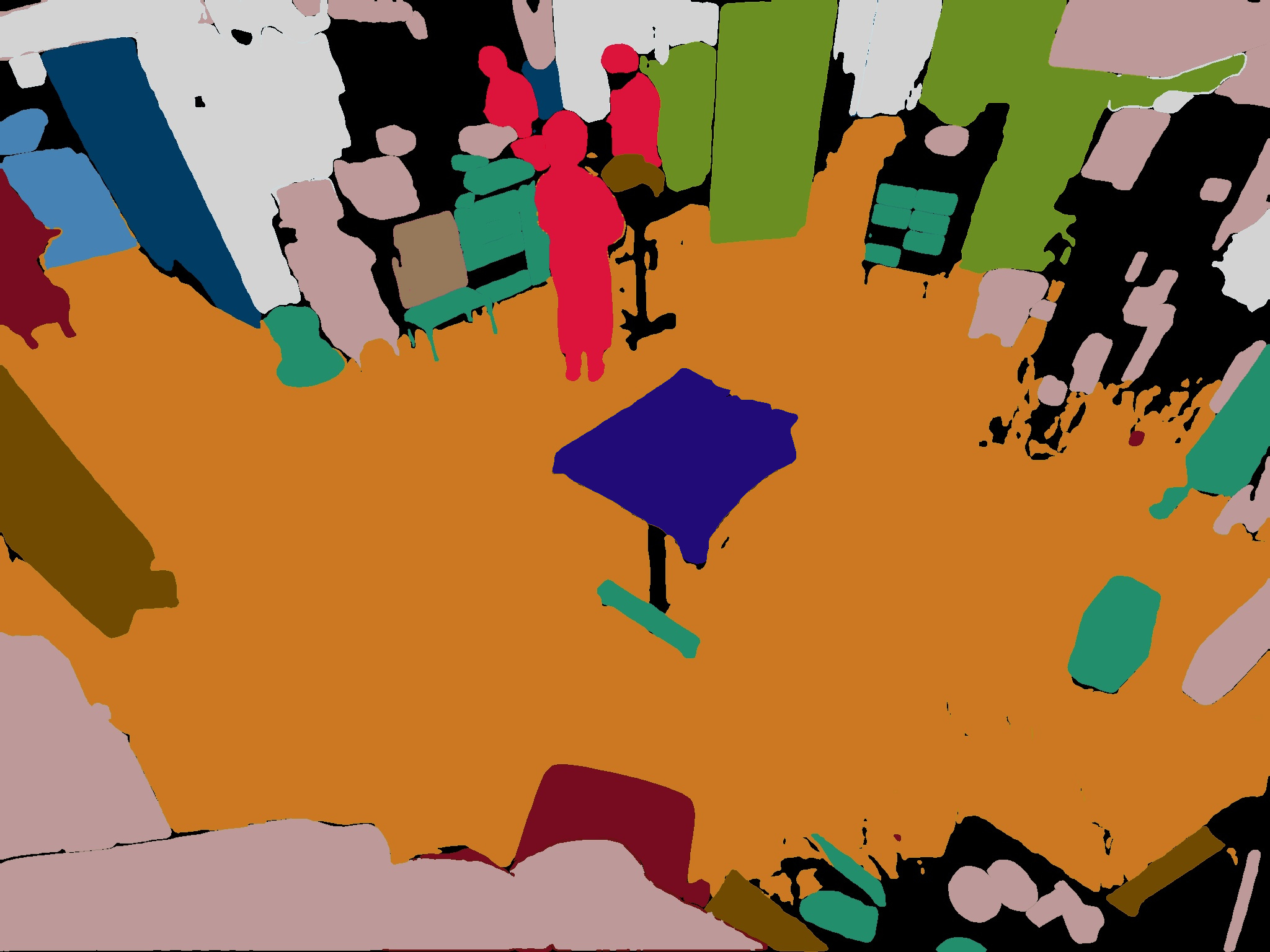}} \\
\\
\vertical{(d) 4D-OR} & 
\raisebox{-0.4\height}{\includegraphics[width=\linewidth]{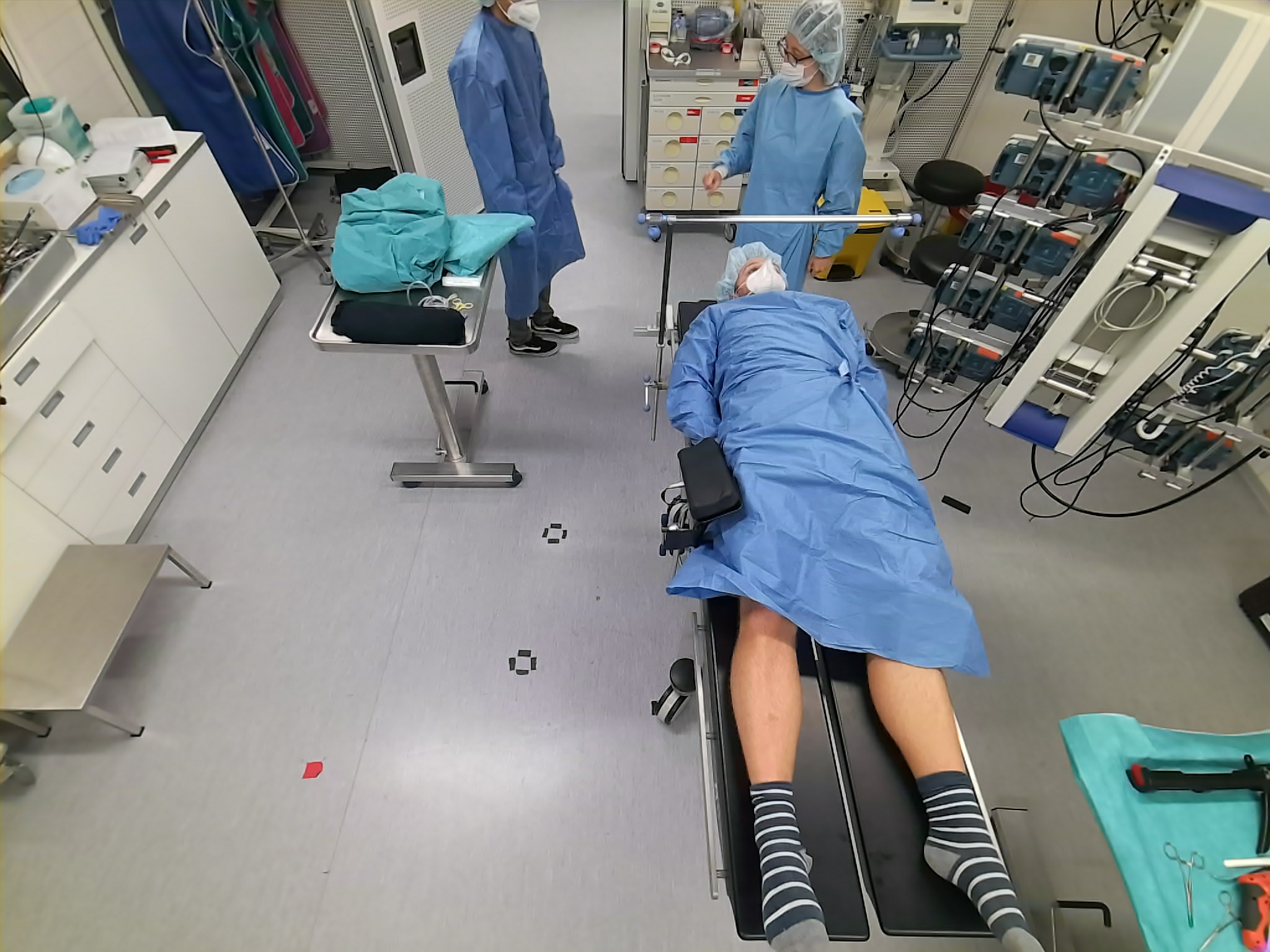}} & \raisebox{-0.4\height}{\includegraphics[width=\linewidth]{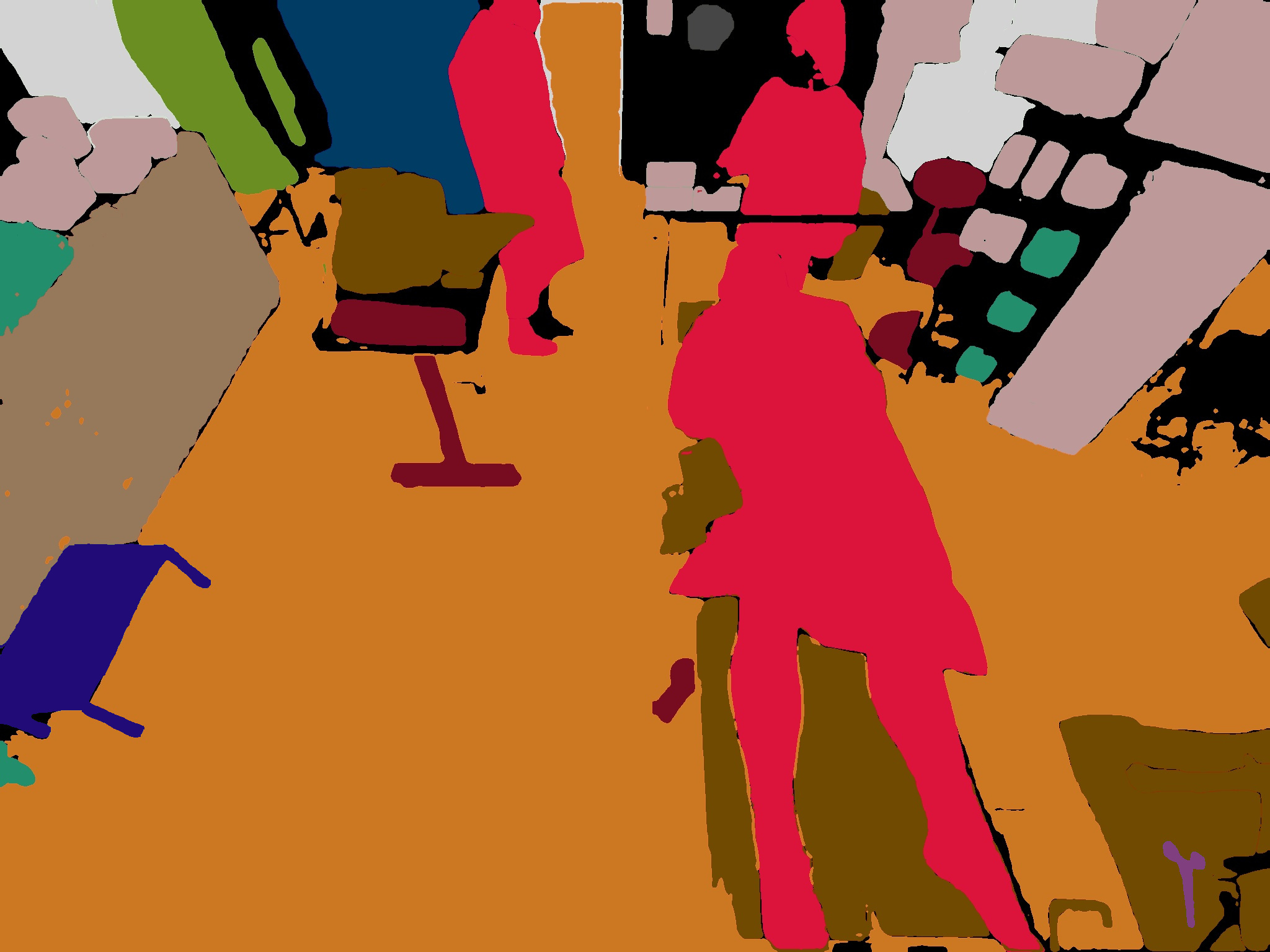}} \\
\\
\multicolumn{3}{c}{\includegraphics[width=0.95\linewidth]{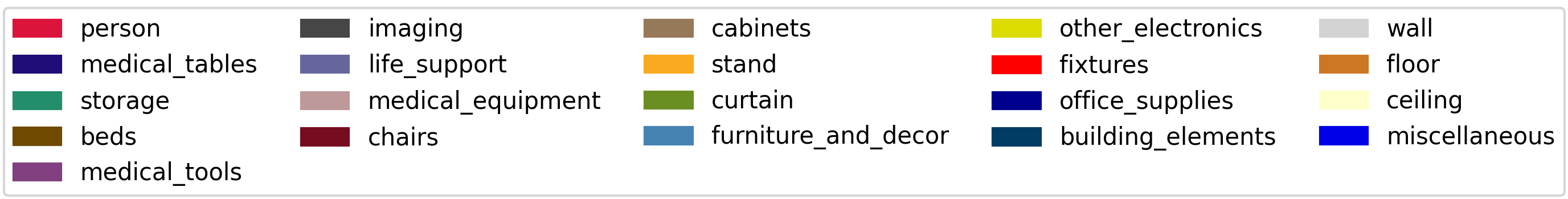}}
\end{tabular}
\caption{Qualitative evaluation of leveraging synthetic knowledge on real-world semantic segmentation within medical environments, featuring images from the MVOR and 4D-OR datasets.}
\label{fig:qualreal}

\end{figure}

\subsection{Generalization to the Real-World}
\label{subsec:ablation}

In this section, we study the generalization of knowledge acquired from our dataset to real-world scenarios, focusing specifically on semantic segmentation. Given the lack of publicly available datasets with annotations analogous to the semantic tasks in \datasetname{}, we present qualitative results. To facilitate this experiment, we developed a naive processing pipeline for real-world images, specifically from the MVOR and 4D-OR datasets. The pipeline starts with a universal image partitioning network to generate class-agnostic object masks. The network that we employ is SAM~\cite{kirillov2023segment}, known for its excellent adaptability to unseen scenarios due to its training on extensive datasets. We produce all possible masks for a given scene using SAM and then use non-maximum suppression to acquire all non-overlapping masks covering the entire scene. Subsequently, we pre-train Segformer on the COCO dataset, the best-performing baseline for semantic segmentation, and then fine-tune it on our dataset. This allows us to generate semantic predictions for the given scene, setting pixel labels to \textit{unlabeled} if the prediction confidence falls below a certain threshold. In the final step, we merge the class-agnostic predictions of SAM with the pixel-wise semantic predictions of Segformer using a majority voting scheme. Specifically, for a given object mask from SAM, we assign the semantic class that appears most frequently in Segformer's predictions. This processing pipeline is completely automatic and no human input is required. \figref{fig:qualreal} presents the qualitative results from this experiment.

We observe that our approach shows encouraging qualitative results. As seen in \figref{fig:qualreal}(a) and \figref{fig:qualreal}(b), the model presents good segmentation of medical equipment, floors, walls, and medical beds, in images from the MVOR dataset. Interestingly, it successfully identifies the socket as fixtures in \figref{fig:qualreal}(a) and accurately predicts a blurred patient's face in \figref{fig:qualreal}(b). However, the approach does present limitations such as overestimation of the medical bed, extending to areas of the display unit's back and the trash can in \figref{fig:qualreal}(a), and misclassification of the doctor's head and patient's hand in \figref{fig:qualreal}(b). In \figref{fig:qualreal}(c) and (d) from the 4D-OR dataset, we observe good segmentation quality for people, chairs, cabinets, storage units, floors, and walls. It is noteworthy that in \figref{fig:qualreal}(d) the scissors are correctly identified as medical tools. Nevertheless, the model falls short in predicting the object on the table in \figref{fig:qualreal}(c), likely due to the NMS step favoring the comprehensive table mask over smaller tool masks. Another point of confusion occurs in \figref{fig:qualreal}(c) where building elements are misclassified as curtains. However, this error is not repeated in \figref{fig:qualreal}(d), suggesting that the model's performance can be influenced by viewing angle or lighting variations. We note that these initial results are promising and show significant potential for future advancements. With continued refinement and the development of more sophisticated methodologies, we foresee substantial progress in exploiting synthetic data for real-world perception tasks in healthcare settings.

\section{Conclusion}

Automatic scene understanding of healthcare facilities is a critical capability for future medical robotics systems. However, the lack of annotated public datasets in these settings has led to slow progress in this domain. This is partly attributable to the privacy considerations involved in the collection of image data in medical environments and the substantial costs associated with annotations. Addressing this concern, we presented \datasetname{}, a multimodal and multitask synthetic dataset for scene understanding of healthcare facilities. The dataset provides a broad set of ground truth annotations for five key perception tasks including object detection, semantic segmentation, instance segmentation, panoptic segmentation, and depth estimation. \datasetname{} consists of over \num{580000} annotations across 31 distinct classes while covering a wide variety of scenes. This diversity comprises 13 different room types, varying illumination conditions, and various types of medical equipment, including surgical robots and tools. With \datasetname{}, we also establish a public benchmark for these tasks that encompasses a range of baselines from fundamental models to state-of-the-art approaches. We believe that this work will pave the way for future innovative research and applications in healthcare settings.


\clearpage
\renewcommand{\baselinestretch}{1}
\setlength{\belowcaptionskip}{0pt}

\begin{strip}
\begin{center}
\vspace{-5ex}
\textbf{\LARGE \bf
Syn-Mediverse: A Multimodal Synthetic Dataset for Intelligent Scene Understanding of Healthcare Facilities} \\
\vspace{2ex}

\Large{\bf- Supplementary Material -}\\
\vspace{0.4cm}
\normalsize{Rohit Mohan, José Arce, Sassan Mokhtar, Daniele Cattaneo, Abhinav Valada}
\end{center}
\end{strip}

\setcounter{section}{0}
\setcounter{equation}{0}
\setcounter{figure}{0}
\setcounter{table}{0}
\setcounter{page}{1}
\makeatletter

\renewcommand{\thesection}{S.\arabic{section}}
\renewcommand{\thesubsection}{S.\arabic{subsection}}
\renewcommand{\thetable}{S.\arabic{table}}
\renewcommand{\thefigure}{S.\arabic{figure}}


\let\thefootnote\relax\footnote{}%

\normalsize

In this supplementary material, we detail the training protocol of all the baselines that we use for benchmarking. We initialize the models with pretrained weights from ImageNet, for each of the segmentation baselines while using the hyperparameters as specified in their original manuscripts unless explicitly stated otherwise.

\section{Object Detection}

{\parskip=5pt\noindent\textit{Faster-RCNN and EfficientDet:} We resize the image within a scale of 0.5 to 2.0. Following, we apply random cropping with dimensions 960$\times$540 along with random flipping. We use Stochastic Gradient Descent for optimization with an initial learning rate of 0.02. We train the models for 120 epochs while reducing the learning rate by a factor of 0.1 at both the 80th and 110th epochs. 

{\parskip=5pt\noindent\textit{YOLO:} We begin by randomly positioning the original image onto a canvas. The size of this canvas falls within a ratio range of 1 to 2 relative to the original image and is populated with the mean values of the RGB channels. Subsequently, we conduct a minimum Intersection-over-Union (IoU) random crop operation utilizing a scale factor of 0.3. The cropped image is subsequently resized to 608$\times$608 dimensions. Additionally, we employ photometric distortion and random flipping as a part of our data augmentation strategy. We use Stochastic Gradient Descent (SGD) with an initial learning rate of 0.001 and train the models for 273 epochs. We then drop the learning rate by a factor of 0.1 at the 218th and then the 246th epoch.  

{\parskip=5pt\noindent\textit{SSD:}We randomly place the original image on a canvas of a ratio within a range of 1 to 4 with respect to the original image filled with the mean of the RGB channels. We then perform a minimum Intersection-over-Union (IoU) random crop with a scale factor of 0.3. The cropped image is then resized to 512$\times$512. We also apply photometric distortion and random flipping for data augmentation. We use Stochastic Gradient Descent with a learning rate of 0.001 for 273 epochs. We then drop the learning rate by a factor of 0.1 at the 218th and then the 246th epoch.  

{\parskip=5pt\noindent\textit{DETR:} We use random flip, resize, and crop as data augmentation. We rescale the image within the range of 0.5 and 2.0. We use 960$\times$540 for the crop size. We employ the AdamW optimizer with a starting learning rate of 0.0001 and train the models for 120 epochs and drop the learning rate by a factor of 0.1 at the 100th epoch.  

\section{Semantic Segmentation} 

We implement a uniform training protocol for all the baselines in terms of data augmentation, optimization, and a learning policy. The images are resized within a ratio range of 0.5 to 2.0. We then employ a random crop with a specified crop size of 960$\times$540, which is subsequently followed by random flipping. For optimization, we utilize Stochastic Gradient Descent with an initial learning rate of 0.001, coupled with a poly learning policy whose power factor is set at 0.9.

\section{Instance Segmentation}

{\parskip=5pt\noindent\textit{Mask-RCNN, DetectorRS, SOLOv2:} We resize the image within a scale of 0.5 to 2.0. We then apply random with dimensions 960$\times$540 along with random flipping. We use Stochastic Gradient Descent for optimization with an initial learning rate of 0.02. We train the models for 120 epochs while reducing the learning rate by a factor of 0.1 at both the 80th and 110th epochs. 

{\parskip=5pt\noindent\textit{YOLOACT:} We randomly place the original image on a canvas of a ratio within a range of 1 to 4 with respect to the original image filled with the mean of the RGB channels. Following this, we perform a minimum Intersection-over-Union (IoU) random crop with a scale factor of 0.3. The cropped image is then resized to 1024$\times$1024.  We also apply photometric distortion and random flipping for data augmentation. We use Stochastic Gradient Descent with a learning rate of 0.001 for 273 epochs. We drop the learning rate by a factor of 0.1 at the 218th and then the 246th epoch.

\section{Panoptic Segmentation}

{\parskip=5pt\noindent\textit{Seamless and EfficientPS:}  We resize the image within a scale of 0.5 to 2.0. We then apply random cropping with dimensions 960$\times$540 along with random flipping. We use Stochastic Gradient Descent for optimization with an initial learning rate of 0.02. We train the models for 120 epochs while reducing the learning rate by a factor of 0.1 at both the 80th and 110th epochs. 

{\parskip=5pt\noindent\textit{Panoptic-Deeplab and Mask2Former:} We use random flipping, resizing, and cropping as data augmentation. We rescale the image within the range of 0.5 and 2.0. We use 960$\times$540 for the crop size. We employ the AdamW optimizer with a starting learning rate of 0.0001 for Mask2Former and employ the Adam optimizer for Panoptic-Deeplab with a starting learning rate of 0.001 and train for 90,000 iterations.

\section{Monocular Depth Estimation}

For all the models, we augment the input images by applying a random rotation within $\pm\ang{2.5}$ and a horizontal flip, both with a probability of 0.5. We then randomly crop the images to a size of $416\times544$ pixels. Finally, we perform a color augmentation with random values of gamma in the $[0.9, 1.1]$ range, brightness within $[0.75, 1.25]$, and color range in $[0.9, 1.1]$ range, after which the RGB values are normalized to the $[0, 1]$ interval.

{\parskip=5pt\noindent\textit{BinsFormer and DepthFormer:} We train both models with a SwinL backbone and a window size of 7, initialized with a pre-training on the NYU dataset, and using the AdamW optimizer for 100 epochs. For BinsFormer, we initialize the learning rate at \num{4e-6}, using a OneCycle schedule, with a maximum learning rate of \num{1e-4}, a warmup period of 8 epochs and a minimum learning rate of \num{4e-8}. For DepthFormer, we use an initial learning rate of \num{1e-4} and a cosine annealing schedule that drops the learning rate to \num{1e-8} over the 100 epochs.

{\parskip=5pt\noindent\textit{SimIPU:} We use a model with a ResNet-50 backbone and a densedepth decoder head. The model is pre-trained in a self-supervised manner using a contrastive loss on the Waymo dataset and then in a supervised manner on the NYU dataset. Once again, we use the AdamW optimizer, with an initial learning rate of \num{4e-6} and train for 100 epochs. We employ a OneCycle schedule, with a maximum learning rate of \num{1e-4}, a warmup period of 8 epochs, and a minimum learning rate of \num{4e-8}.

\end{document}